\documentclass{article}

\PassOptionsToPackage{numbers, compress}{natbib}

\usepackage[final]{neurips_data_2023}

\usepackage[utf8]{inputenc} 
\usepackage[T1]{fontenc}    
\usepackage{hyperref}       
\usepackage{url}            
\usepackage{booktabs}       
\usepackage{amsfonts}       
\usepackage{nicefrac}       
\usepackage{microtype}      
\usepackage[dvipsnames]{xcolor}         
\usepackage{epsfig}
\usepackage{graphicx}
\usepackage{amsmath}
\usepackage{amssymb}
\usepackage{booktabs}
\usepackage{xspace}
\usepackage{enumitem}
\usepackage{csquotes}

\def\cpar{\hss\egroup\line\bgroup\hss}

\makeatletter
\DeclareRobustCommand\onedot{\futurelet\@let@token\@onedot}
\def\@onedot{\ifx\@let@token.\else.\null\fi\xspace}
\def\eg{\emph{e.g}\onedot} 
\def\ie{\emph{i.e}\onedot}

\makeatother

\title{CHAMMI: A benchmark for channel-adaptive models in microscopy imaging}

\author{%
  Zitong S. Chen\thanks{These authors contributed equally to this work} \\
  Broad Institute\\
  Cambridge, MA \\
  \And
  Chau Pham\footnotemark[1] \\
  Boston University \\
  Boston, MA \\
  \And
  Siqi Wang \\
  Boston University \\
  Boston, MA \\
  \And
  Michael Doron \\
  Broad Institute\\
  Cambridge, MA \\
  \And
  Nikita Moshkov \\
  Biological Research Centre \\
  Szeged, Hungary \\
  \And
  Bryan A. Plummer \\
  Boston University \\
  Boston, MA \\
  \And
  Juan C. Caicedo \\
  Univ. of Wisconsin-Madison \\
  Madison, WI \\
}

\begin{document}

\maketitle

\begin{abstract}
\setcounter{footnote}{0} 
Most neural networks assume that input images have a fixed number of channels (three for RGB images). However, there are many settings where the number of channels may vary, such as microscopy images where the number of channels changes depending on instruments and experimental goals. Yet, there has not been a systemic attempt to create and evaluate neural networks that are invariant to the number and type of channels. As a result, trained models remain specific to individual studies and are hardly reusable for other microscopy settings. In this paper, we present a benchmark for investigating channel-adaptive models in microscopy imaging, which consists of 1) a dataset of varied-channel single-cell images, and 2) a biologically relevant evaluation framework. In addition, we adapted several existing techniques to create channel-adaptive models and compared their performance on this benchmark to fixed-channel, baseline models. We find that channel-adaptive models can generalize better to out-of-domain tasks and can be computationally efficient. We contribute a \href{https://doi.org/10.5281/zenodo.7988357}{curated dataset}\footnote{https://doi.org/10.5281/zenodo.7988357} and an \href{https://github.com/broadinstitute/MorphEm.git}{evaluation API}\footnote{https://github.com/broadinstitute/MorphEm.git} to facilitate objective comparisons in future research and applications.

\end{abstract}

\section{Introduction}

Microscopy images routinely enable a myriad of applications in experimental biology, including tracking living cells, characterizing diseases, and estimating the effectiveness of treatments. The analysis of microscopy images often requires quantitative methods that capture differences between cellular conditions, e.g., differences between control and treated cells in a vaccine trial. To accomplish effective cell-morphology quantification, deep learning has been adopted in problems such as cell segmentation \cite{stringer2021cellpose}, sub-cellular protein localization \cite{le2022analysis}, and compound bioactivity prediction \cite{moshkov2023predicting}. 

Unlike natural images (stored in RGB format), microscopy images are often multiplexed and can span a variety of specialized colors in different channels \cite{bray2016cell, goltsev2018deep, saka2019immuno}. The number of channels in a microscopy study is a choice made by biologists depending on various factors that include instrument capabilities and experimental needs. There is no universal standard for acquiring images with a fixed set of channels, and instead, novel imaging techniques push the limits with more and more fluorescent markers measured simultaneously \cite{hickey2022spatial}. This flexibility increases the potential to observe specific biological events, but poses practical challenges for established computer vision methods. For example, having varying channels limits the ability to reuse pre-trained models from one study to another. Moreover, training models that are specific to one microscopy configuration may result in suboptimal performance because of lack of sufficient data, and may also result in high risk of learning spurious correlations rather than useful biological features \cite{scholkopf2021toward}.

Can computer vision models in microscopy process a flexible number of input channels? In principle, there is no technical limitation to creating models that adapt to a varying channel dimension, although some inspiration may come from models designed to process varying-size sequences and sets \cite{mikolov2010recurrent, zaheer2017deep, vaswani2017attention}. However, there is no systematic attempt to design neural networks that are agnostic to the number of channels in an image, and it remains unclear whether such models would bring any benefits over training specialized, fixed-channels models. We hypothesize that creating models that accommodate multi-channel images can improve microscopy image analysis by: 1) saving the time and resources of training models for new imaging configurations; 2) accelerating the pace of biological research with reusable pre-trained channel adaptive models; 3) improving performance in small datasets with transfer learning. We envision channel-adaptive models trained at large scale using many datasets that may seem incompatible at first, but that after pooling them together, can unlock the potential for higher level performance and other emerging behaviors in microscopy image analysis. 

In this paper, we present a benchmark for exploring CHannel-Adaptive Models in Microscopy Imaging (CHAMMI). Our goal is to facilitate the development and evaluation of machine learning models and architectures that can adaptively process microscopy images from different studies. Our benchmark has two main components: 1) a dataset of single-cell images collected from three high-profile, publicly available biological studies. We combined these image sources into a channel-varying dataset. 2) An evaluation framework and metrics for nine downstream tasks to assess performance on biologically relevant problems, which enables comparison of methods under different settings including out-of-distribution generalization tests. In addition, we present an experimental evaluation of baseline methods and existing techniques that we adapted to solve the varying channels problem. We find that there is ample room for improvement in terms of algorithms and architectures for creating channel-adaptive models that can be of general use in microscopy imaging.

\begin{figure}
  \centering
  \includegraphics[width=\linewidth]{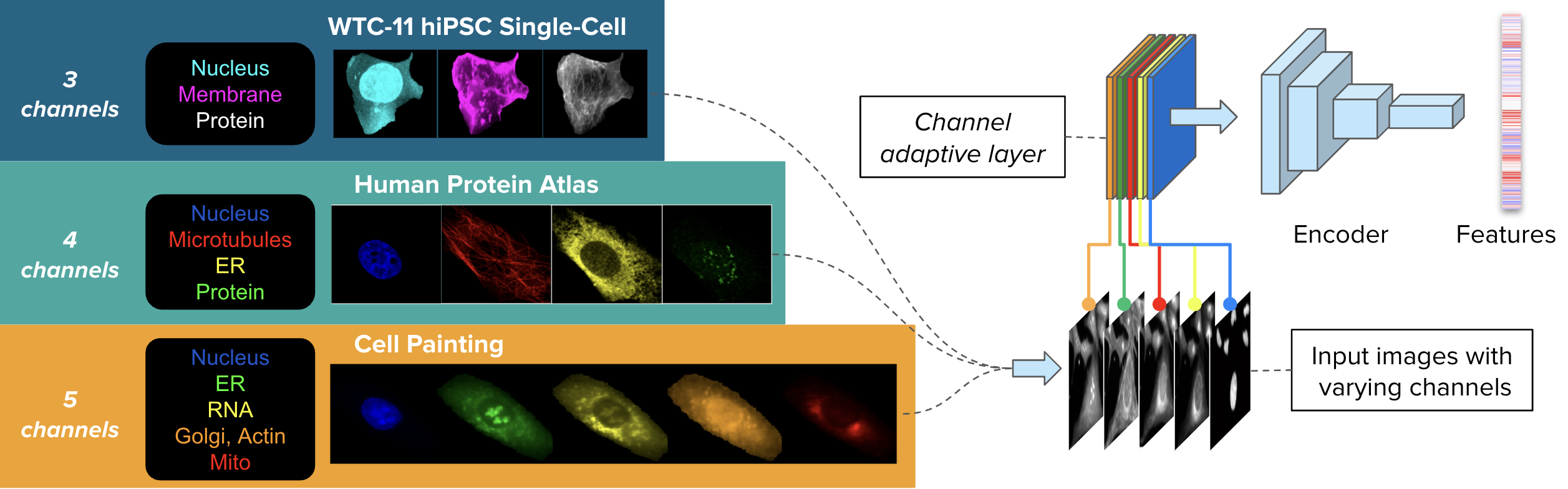}
  \caption{\small{Example CHAMMI images (left) and illustration of a channel-adaptive model (right). The dataset consists of varying-channel images from three sources: WTC-11 hiPSC dataset (WTC-11, 3 channels), Human Protein Atlas (HPA, 4 channels), and Cell Painting datasets (CP, 5 channels). The model takes images of cells with varying number of channels and produces feature embeddings for downstream biological applications.}}
  \label{fig:example}
\end{figure}

\section{Related Work}

Researchers have extensively explored a wide range of architectures and techniques to better represent images, \eg, VGGNet~\citep{vgg}, ResNe(X)t~\citep{resnet1,resnet2}, MobileNet~\citep{mobilenet}, EfficientNet~\citep{efficientnet}, ViT~\citep{vit}, SWIN~\citep{swin}, ConvNeXt~\citep{liu2022convnet}, and Hiera~\citep{hiera}, among others. However, these models have been developed to operate specifically with RGB images. Thus, researchers exploring different imaging modalities, such as spectral images, RGB-D (depth) images, thermal images, and ultrasound images~\citep{huang2022spectral,zhang2022overview,bhattacharyya2021deep}, manually modify these models to handle a fixed, but different number of channels. Additionally, these solutions typically train a new, separate model for each imaging modality, whereas in this work we aim to train a single model that supports images with varying channels.

Our goal is similar in spirit to methods that add new dimensions to existing image encoders. For example, data with temporal or volumetric dimensions, such as videos, and 3D volumes (\eg, Multi-view Images, Point clouds)~\citep{wang2023all,chen2023hnerv,aoki2019pointnetlk}. 
However, these methods still assume that a single frame or 3D slice contains a set number of channels (\eg, each frame in a video being an RGB image), whereas images in our setting can vary in the number and type of available channels.

Image-based profiling and representation learning have emerged as a crucial component in the study of cellular morphology. Various deep learning models have been proposed to learn representations that capture the patterns in cellular morphology~\citep{li2021cell,zhang2021neuron,ceran2022tntdetect,contreras2022machine,moshkov2022learning}, and span multiple applications, including cell-line prediction~\citep{yao2019cell,doan2020objective},  protein localization ~\citep{kobayashi2022self}, and drug discovery~\citep{chandrasekaran2021image}. Recent interest has emerged in generalist models for microscopy that are not dataset specific, such as using ImageNet pre-trained networks \citep{ando2017improving}. CytoImageNet \citep{hua2021cytoimagenet} pioneered the creation of a diverse microscopy image dataset for representation learning, and Microsnoop \citep{xun2023microsnoop} created self-supervised models at large scale to profile any type of microscopy image. While these efforts do not investigate channel-adaptive models, they demonstrate the potential for dataset-agnostic approaches in biological applications.
\section{CHAMMI}
\label{sec:CHAMMI}

The CHAMMI (channel-adaptive models in microscopy imaging) benchmark comprises a single-cell image dataset curated from three publicly available resources: the Allen Institute Cell Explorer, the Human Protein Atlas, and the Cell Painting Gallery (Appendix). These have images with three, four, and five channels, respectively. We collected and standardized single-cell images to have comparable resolution to facilitate the development of channel-adaptive models, and we designed nine downstream tasks to evaluate performance. 

\begin{figure}
  \centering
  \includegraphics[width=\linewidth]{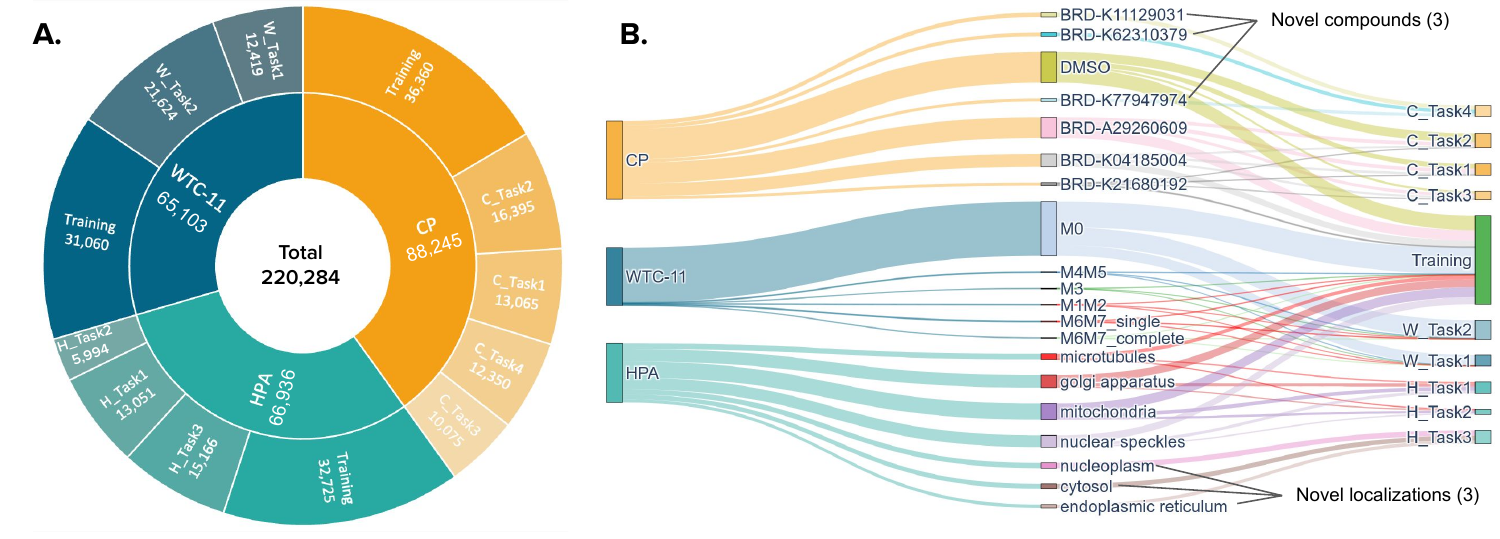}
  \caption{Summary statistics of the CHAMMI dataset. A) Number of images from each source split by training and testing sets. The training set has images from all three sources, whereas the validation sets are specific to one source. B) Distribution of images with various classification labels across training and testing sets. Each image is annotated with one of six (WTC-11) or seven (HPA, CP) labels. WTC-11 images are labeled by the cell-cycle stage of single cells, HPA images are labeled by the protein subcellular localization, and the CP images are labeled by the compound treatment.}
  \label{fig:data-stats}
\end{figure}

\subsection{Microscopy Image Datasets}

\textbf{The WTC-11 dataset}: this is a collection of more than 200,000 single cell images in 3D and at high-resolution~\citep{viana2023integrated}. The dataset was created with cells tagged with a fluorescent marker for one of 25 cellular structures or major organelles. Each image has three channels: cell membrane, the nucleus, and the organelle of interest. We used the max-projection of the 3D volume of each channel into a 2D plane, rendering it a conventional three-channel image with 238x374 pixels. The original dataset was collected to study cellular variation in normal stem cells, and provides a variety of biological annotations. In our benchmark, we include annotations associated with cell-cycle stages, which are organized into six categories with a sample of 65,103 images from six organelles. 

\textbf{The HPA dataset}: a collection of more than 80,000 images containing multiple single cells, and stained with fluorescent markers for four cellular structures: microtubules, nucleus, endoplasmatic reticulum and a protein of interest~\citep{thul2017subcellular, ouyang2019analysis, le2022analysis}. The dataset has images of about 30 human cell lines, and covers more than 13,000 proteins to study their localization patterns within cells. Each image is annotated with a few manual labels that indicate the localizations of the visible protein; there are 20 localization classes in total. We used a segmentation model provided by the dataset creators to get single-cell images \citep{le2022analysis}. Then, a sample of 66,936 single-cell images from 18 cell lines and 8 protein localization classes were included in our benchmark with their corresponding annotations.

\textbf{The Cell Painting dataset}: a set of $\sim$8M single-cell images collected from five perturbation studies to train machine learning models~\citep{way2022morphology,bray2017dataset,gustafsdottir2013multiplex}. These studies used the Cell Painting assay, which stains eight cellular compartments with six fluorescent markers imaged in five channels. The goal of these studies was to quantify the response of cells to different treatments or perturbations, which is an essential evaluation in drug discovery projects. This dataset includes perturbations with 400 compounds and 80 gene over-expression experiments tested in two cell lines with four or five replicates. We sampled 88,245 single-cell images from seven compound experiments, including the negative control. 

\subsection{Prediction tasks and evaluation}

\begin{figure}
  \centering
  \includegraphics[width=\linewidth]{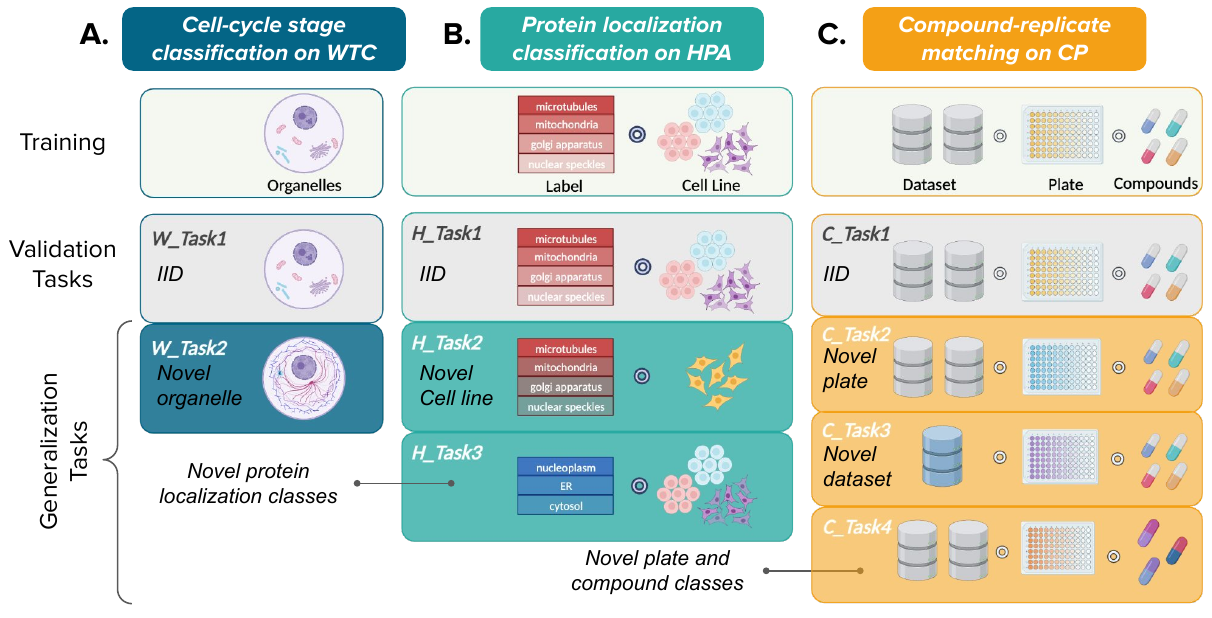}
  \caption{Illustration of the evaluation tasks in CHAMMI with training, validation (gray) and generalization (colored) tasks. A) Cell-cycle stage classification on WTC-11 (6 classes), organized in two tasks stratified by the organelle observable in the protein channel. B) Protein localization classification on HPA (7 classes), organized in three tasks stratified by class labels and cell lines. C) Compound replicate matching on Cell Painting (7 compounds), organized in four tasks stratified by source dataset, plate ID, and compound treatment. \emph{IID}: independent and identically distributed.}
  \label{fig:tasks}
\end{figure}

We present nine tasks with increasing levels of complexity to evaluate the ability of models to generalize to new biologically-relevant experimental regimes (Fig.~\ref{fig:tasks}). The tasks are grouped into validation and generalization tasks (Fig.~\ref{fig:tasks}). Validation tasks (suffix = 1 in the ID) are classification problems where the test data follows the same distribution as the training data (IID). Generalization tasks (suffix $>1$ in the ID) are problems with out-of-distribution (OOD) test data. Generalization tasks split the datasets by leaving certain biological samples out, increasing the difficulty and reflecting real-world application conditions. Image sampling was stratified such that training and testing images have the same distribution in terms of biological labels and technical variability.

\textbf{Cell-cycle stage classification on the WTC-11 dataset}: since many disease-inducing disruptions occur during mitosis \cite{nurse1998understanding}, precisely determining the cell-cycle stage of cells in high-throughput screens can improve therapeutic development. Here, the goal is to classify single-cell images into one of six categories (appendix). The 65,103 single-cell images included in CHAMMI are split into one training set and two test sets. The training set contains cells with one of four cellular compartments fluorescently tagged: nuclear speckles, mitochondria, microtubules, or Golgi apparatus. 

\begin{itemize}[nosep,leftmargin=*]
    \item \textbf{W\_Task1} is an IID validation task and contains 12,419 test images fluorescently tagged with the same set of cellular compartments as the training set. 
    \item \textbf{W\_Task2} evaluates the ability of models to predict cell-cycle stages when a different cellular structure is fluorescently tagged. It contains 21,624 test images tagged with a new set of three cellular compartments: tight junctions, centrioles, or actin bundles. 
\end{itemize}

\textbf{Protein localization classification on the HPA dataset}: accurate protein localization prediction is of great interest to clinical research because their mislocalization can lead to numerous diseases, including cancer~\citep{wang2014protein}. Note that protein localization is actually a classification problem with seven categories: 1) nuclear speckles, 2) mitochondria, 3) microtubules, 4) Golgi apparatus, 5) nucleoplasm, 6) cytosol, and 7) endoplasmic reticulum (ER). In CHAMMI, we included 66,936 single-cell images sampled from the original HPA dataset. Since a protein can have multiple localization annotations, we filtered for images with only one annotation to avoid ambiguity. The training set includes cells with one of four localizations (1,2,3,4), and the cells come from one of 17 cell lines.

\begin{itemize}[nosep,leftmargin=*]
    \item \textbf{H\_Task1} is an IID task with 13,051 single-cell images in four classes and 17 cell lines.
    \item \textbf{H\_Task2} evaluates protein localization prediction in a novel cell line: HEK-293, which is unseen during training. This task contains 5,994 images and class labels are the same as in the training set.
    \item \textbf{H\_Task3} tests generalization with a novel set of class labels not present in the training set (5,6,7). It has 15,166 images from the same set of 17 cell lines as in the training set.
\end{itemize}

\textbf{Compound replicate matching on Cell Painting datasets}: in drug development and repurposing pipelines, compound-replicate matching is a standard test used to study the reproducibility of perturbation experiments \cite{caicedo2017data}. Ideally, the features of cells should consistently match other cells treated with the same perturbations, and ignore batch effects caused by non-biological or technical artifacts. We selected 6 compound perturbations and one negative control from the source datasets (appendix). The goal is to predict which perturbation was introduced to cells by matching images using nearest-neighbor (NN) search. The training set includes cells perturbed with three treatments and the DMSO negative control. In total, CHAMMI includes 88,245 cell images sampled from 10 plates and two publicly available Cell Painting datasets: LINCS and CDRP. 

\begin{itemize}[nosep,leftmargin=*]
    \item \textbf{C\_Task1} is an IID task with 13,065 images in 4 classes, taken from 9 plates and 2 source datasets. 
    \item \textbf{C\_Task2} images come from the same set of two datasets but a novel set of 3 plates, with the same set of four perturbations as in the training set.
    \item \textbf{C\_Task3} evaluates a model's ability to generalize to a novel data source (BBBC022), with 10,075 cells from a new set of four plates. The perturbation labels are the same as in the training set.
    \item \textbf{C\_Task4} evaluates generalization ability to novel treatments in the same set of two source datasets as in the training set. This set includes 12,350 cells.
\end{itemize}

\textbf{Evaluation Procedure.} Retrieval and clustering tasks are typical in cellular data analyses because biological research aims to discover unknown phenomena. Usually, the goal is to reveal differences and similarities among cellular phenotypes instead of using pre-trained classifiers to correctly predict known category labels. For this reason, we evaluate all tasks using a nearest-neighbor (NN) search approach based on feature matching with the cosine similarity. In the presence of ground truth annotations, this reduces the evaluation procedure to a classification problem with 1-NN search. 

NN search can resolve predictions for test data in IID tasks (W\_Task1, H\_Task1, C\_Task1) using only the training set as a reference. Also, OOD tasks that share labels with the training set but have novel image features can be resolved in the same way (W\_Task2, H\_Task2, C\_Task2, C\_Task3). However, some OOD tasks introduce novel class labels not available in the training set. In this case, we combine the test data with the training set to allow the NN search procedure to resolve the corresponding predictions (H\_Task3, C\_Task4). For these cases, we hide the labels of one test data point at a time (leave-one-out). Note that this still keeps the test data out of training deep learning models to prevent learning anything from these images, and to keep the hold-out set private for evaluation only.

\textbf{Evaluation Metrics.} As a performance metric, we use the macro-average F1-score of predictions with a 1-NN classifier for each task independently. We also define the \emph{CHAMMI Performance Score} (CPS) as the weighted average of the six generalization tasks as follows: $\text{CPS} = W_{Task\_2}/3 + \left(H_{Task\_2} + H_{Task\_3}\right)/6 + \left(C_{Task\_2} + C_{Task\_3} + C_{Task\_4}\right)/9$, where $D_{Task\_i}$ is the F1-score of task $i$ in dataset $D$. See the supplementary for more details.

\newcommand{\ChRep}{{\fontfamily{lmss}\selectfont ChannelReplication}}
\newcommand{\FxChn}{{\fontfamily{lmss}\selectfont FixedChannels}}
\newcommand{\Depth}{{\fontfamily{lmss}\selectfont Depthwise}}
\newcommand{\SlPrm}{{\fontfamily{lmss}\selectfont SliceParam}}
\newcommand{\TgPrm}{{\fontfamily{lmss}\selectfont TargetParam}}
\newcommand{\TmMix}{{\fontfamily{lmss}\selectfont TemplateMixing}}
\newcommand{\HpNet}{{\fontfamily{lmss}\selectfont HyperNet}}

\section{Experiments and Results}
\label{sec:results}

\begin{figure}
  \centering
  \includegraphics[width=1.0\linewidth]{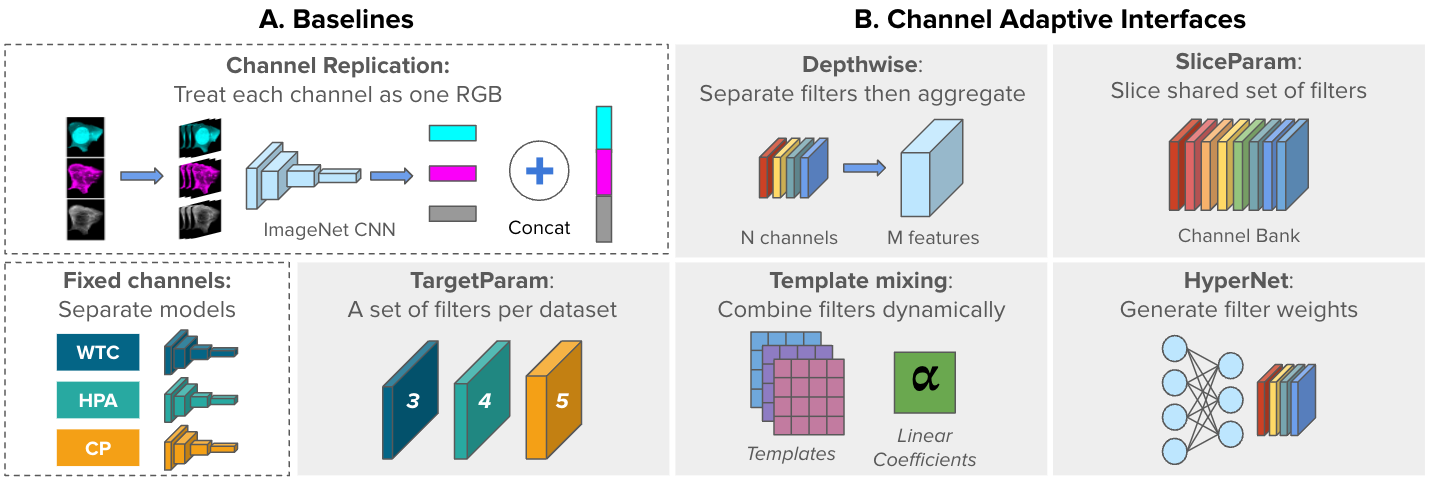}
  \caption{Illustration of the evaluated models. A) Two non-adaptive, baseline approaches: \ChRep~and \FxChn. B) Five channel-adaptive strategies to accommodate varying image inputs: \Depth, \SlPrm, \TgPrm, \TmMix, and \HpNet~(gray blocks). Adaptive interfaces are the first layer of a shared backbone network. Descriptions are provided in Sec.~\ref{sec:architectures} and the supplementary contains additional details.
}
  \label{fig:diagrams}
\end{figure}

We investigate three main aspects of channel-adaptive models that can be evaluated with the proposed CHAMMI benchmark: 1) channel-adaptive architectures; 2) training considerations, and 3) generalization capabilities. We conducted all experiments using a ConvNeXt model \citep{liu2022convnet}, which is an efficient and accurate convolutional architecture for natural image classification. When using pre-trained weights, the model is pre-trained on ImageNet 22K~\citep{deng2009imagenet}. For channel-adaptive models, we replace the first layer of the ConvNeXt model and keep all the remaining layers as they are. 
More details on the implementation and experimental setup can be found in the supplementary.

\subsection{Channel-adaptive architectures}
\label{sec:architectures}

\textbf{Non-adaptive baselines}. We consider baseline  strategies that require minimal adaptations from existing computer vision architectures  (Fig.~\ref{fig:diagrams}A). The first is \emph{\ChRep}, a common transfer-learning strategy for profiling cellular morphology~\cite{liberali2015single, pawlowski2016automating}. \ChRep~reuses an ImageNet pre-trained model for extracting features from each channel independently, followed by concatenation of feature vectors. This has the advantage of not requiring any further training, but it leads to a higher computing cost and feature dimensionality that is proportional to the number of channels. \ChRep~ sets the baseline with a performance score of 0.344. The second baseline is \emph{\FxChn}, which sets the number of input channels in the model as needed and trains a separate model for each variation. This results in three separate models to solve the tasks in CHAMMI. Trained \FxChn~models obtain a performance score of 0.500, improving 45\% relative to \ChRep, with higher scores in almost all individual tasks (Fig.~\ref{fig:baselines}A). This confirms that ImageNet pre-trained weights extract features that, out of the box, are not optimal for microscopy images, as they do not capture relevant cellular morphology from each channel on their own nor when used jointly over the available channels.

\textbf{Channel-adaptive strategies}. We propose three simple extensions of convolutional networks to create channel-adaptive models (Fig. \ref{fig:diagrams}B), and train them with varying-channel images in CHAMMI with randomly initialized weights. The first is \emph{\Depth~}convolution, which applies a single filter for each of the channels. Filter responses are then combined and reduced to a feature representation by averaging across channels. A trained \Depth~model achieves a performance score of 0.517, generally improving upon \FxChn~models. 
The second model, \emph{\SlPrm}, consists of a set of convolutional filters for each channel that are dynamically allocated according to the needs of the input image. Concretely, given an input image with $N$ channels, \SlPrm~slices the $N$ corresponding sets of filters to form a new convolutional layer for each set of channels.  \SlPrm~achieves a performance score of 0.457, better than \ChRep, but behind \FxChn. The third extension is \emph{\TgPrm}, which trains a separate convolutional layer for each set of channels and has shared backbone weights for all images. A trained \TgPrm~model achieves a 0.496 score, improving performance in some tasks (Fig.~\ref{fig:baselines}B). 

We also propose adaptations of two existing strategies for approaching the channel-adaptive model problem (Fig.~\ref{fig:diagrams}B). First, \emph{\TmMix}~\citep{savarese2018learning,plummerNPAS2022}, which learns a linear combination of shared parameter templates to generate the weights for a layer. On the one hand, this mechanism enables the information to traverse across channels as they share the same templates. On the other hand, by representing layer weights as a linear combination of these shared templates, each channel can preserve its unique features and patterns. \TmMix~achieves a performance score of 0.466. The other approach is \emph{\HpNet}~\citep{haHypernetworks2016}, which uses a simple MLP to generate layer's weights for other models. For our setting, we generate the weights for each channel independently and then concatenate them together. \HpNet~ achieves a performance score of 0.537, showing consistent improvements in all CHAMMI tasks (Fig.~\ref{fig:baselines}B).

Channel-adaptive models exhibit improved capacity to perform well in the CHAMMI tasks. When trained from scratch with all varying-channel images in the training set, \Depth, \TgPrm, and \HpNet~ approach the same level of performance of the corresponding \FxChn~baseline (Fig.~\ref{fig:baselines}B), and even improve in some of the tasks. In this regime, \HpNet~was able to improve performance in all nine tasks, supporting the idea that a single, channel-adaptive model can be used to conduct microscopy image analysis across datasets with different number of channels, both computationally efficiently and accurately. This validates our hypothesis that sharing layers in a network across datasets that have varying numbers of channels can boost performance over training completely separate models for every dataset.

\subsection{Training and optimization}

\begin{figure}
  \centering
  \includegraphics[width=\linewidth]{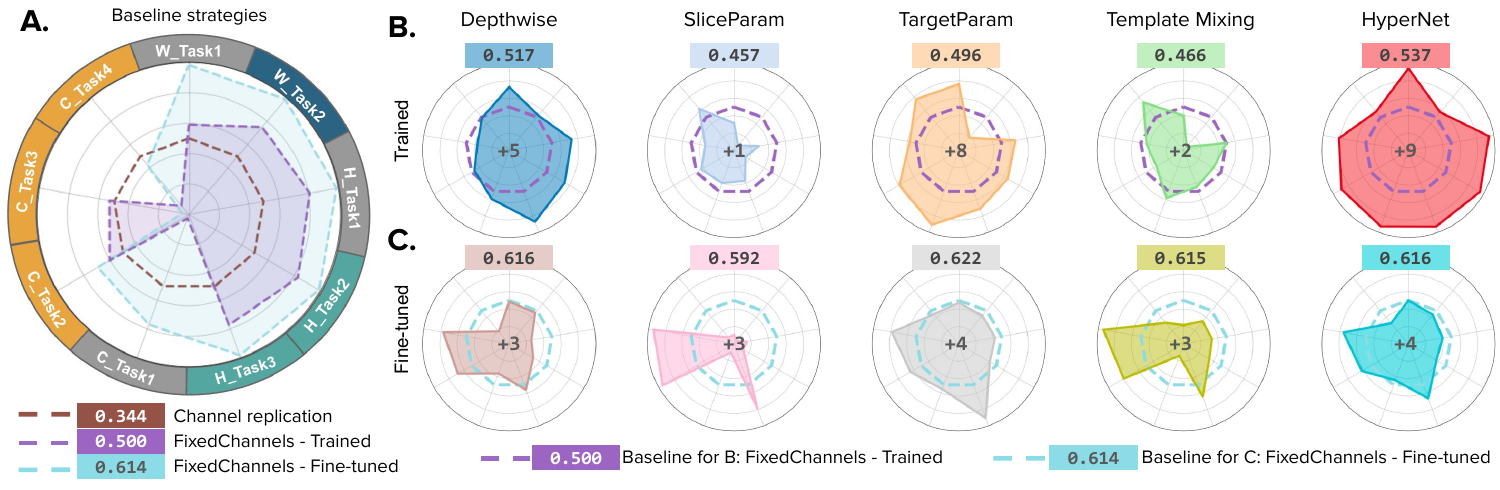}
  \caption{Model comparison on the CHAMMI benchmark. Radar plots have nine axes representing the benchmark tasks. Numbers in color boxes indicate the performance score obtained by the model. A) Comparison of three baseline models. The tasks in the outer circle are colored by the dataset they correspond to, with gray meaning validation tasks, and colors meaning generalization tasks. B,C) Comparison of five, channel-adaptive strategies trained or fine-tuned with the CHAMMI dataset. Number in the circle center indicates in how many tasks the model is better than the baseline.
}
  \label{fig:baselines}
\end{figure}

\textbf{Training vs Fine-tuning}. \FxChn~and channel-adaptive models can be \emph{trained from scratch} by initializing weights randomly, or \emph{fine-tuned} by initializing with ImageNet pre-trained weights. For fine-tuning, the weights of the first layer are replicated to cover more than three channels as needed. \FxChn-fine-tuned models improve performance about 22\% over randomly initialized models from 0.500 to 0.612 in the performance score. Note some tasks that under-perform after fine-tuning: generalization tasks C\_Task3 and C\_Task4 in the Cell Painting data (Fig.~\ref{fig:baselines}A). This reveals how \FxChn~models overfit to variation specific to one type of image, reducing a model's effectiveness on OOD problems. While fine-tuned \FxChn~models are a higher bar to improve, all channel-adaptive strategies in this regime are able to improve the performance of generalization tasks H\_Task3, C\_Task2, and C\_Task3 (peaks outside the cyan polygon in Fig.~\ref{fig:baselines}C). This highlights the benefits of using channel-varying data instead of fine-tuning for fixed subsets. 

Most channel-adaptive strategies outperform the \FxChn~baseline when fine-tuning, indicating that generalist models can have better performance. On average, fine-tuning channel-adaptive models improves performance by 23\% relative to training from scratch (Tab.~\ref{tab:cost}).  The large difference between training and fine-tuning models indicates that CHAMMI still benefits from pre-training on larger datasets, even if they are not cellular images. CHAMMI is a small dataset with well balanced, and well annotated images useful for fine-tuning and evaluation rather than for pre-training at large scale. Our future work includes the extension and maintenance of CHAMMI with a large-scale non-annotated dataset that is useful for pre-training.

\textbf{Data augmentation and loss functions}. Common data augmentations for RGB images are not necessarily effective for microscopy because of differences in color space. We implemented a set of basic augmentations that have been found to be successful for fluorescence microscopy, including random cropping and flips. In addition, we evaluated the performance of thin-plate-spline (TPS) transformations \citep{donato2002approximate} as a potential strategy to simulate technical artifacts and improve model robustness to noise. We observed that more complex data augmentations can increase performance in all architectures up to 2\% relative to basic augmentations (Tab.~\ref{tab:cost}), highlighting the importance of image manipulation for OOD generalization \citep{huang2022harnessing,xie2020unsupervised,kang2023improving,zhang2020generalizing, pernice2023out}.

We also investigated self-supervised learning (SSL) \citep{chen2020simple, he2020momentum, grill2020bootstrap, jiang2023hierarchical,caron2021emerging,oquab2023dinov2} for solving the tasks in the CHAMMI benchmark. The results in Fig. \ref{fig:baselines} were obtained with a supervised loss using class annotations in the CHAMMI dataset. Here, we evaluate SimCLR \citep{chen2020simple,chen2020big} using the basic set of augmentations plus TPS transformations \citep{donato2002approximate}. We observe that SimCLR alone severely underperforms (SSL \FxChn~performance score is 0.360 vs 0.623 when using supervision). However, in combination with supervised learning, SimCLR improves performance by about 2\% (\FxChn~supervision+SSL performance score is 0.633, Tab.~\ref{tab:cost}). These results reinforce our observation that CHAMMI is a great dataset for fine-tuning and benchmarking, but not necessarily for large-scale pre-training with SSL.

\textbf{Computational cost}. Training a channel-adaptive model not only has the potential to improve performance in downstream tasks, but can also be computationally more efficient, reducing the carbon footprint of microscopy image analysis at scale. We estimated the computational cost of models both for training and testing times, which are reported in Tab.~\ref{tab:cost}. The training cost is estimated with the number of parameters of the underlying architecture, assuming that the specific training strategy adds a constant cost to the overall training complexity. The inference cost is estimated with the average number of forward passes required by a model to produce features for a single-cell image in the CHAMMI test sets. Note that \ChRep~has the lowest training cost, but the highest inference cost, because it needs to process channels individually. 

The training cost of channel-adaptive models is very similar to each other, because all are based on the same ConvNext architecture, which is trained once for all channel configurations. The differences are primarily in the arrangement of the first convolutional layer, which may require more or less filters depending on the strategy. Inference cost is also the largely same as all models process each single-cell image once. Both TemplateMixing~\citep{savarese2018learning,plummerNPAS2022} and HyperNet~\citep{haHypernetworks2016} have a computational overhead due their layer weight generation requirements.  However, this increase is negligible  when amortized across a batch of images (\eg, less than 0.2\% for a batch size of 64)~\citep{plummerNPAS2022}, and needs to only be computed once when processing multiple batches of images with the same set of channels. Thus, the main difference between models is on performance accuracy, which is reported in Tab.~\ref{tab:cost} as the overall score (performance on the six generalization tasks). Most models maintain the same level of performance or improve upon the \FxChn~baselines.

\begin{table}
  \caption{Computational cost and performance score for each method on CHAMMI. The cost is the number of trainable parameters and the average number of forward passes during inference. Performance scores include relative improvement percentages with respect to the previous column. From left to right: Trained, Fine-tuned, Fine-tuned + TPS, Fine-tuned + TPS + SSL. }
  \label{tab:cost}
  \centering
  \begin{tabular}{lrccccccc}
    \toprule
    & \multicolumn{2}{c}{Computational Cost $\downarrow$}  & \multicolumn{4}{c}{Performance Score $\uparrow$}    \\
    \cmidrule(r){2-3} \cmidrule(r){4-7}
    Model     & \textit{N}\textsubscript{params}   & \textit{C}\textsubscript{inference} &  Trained & Fine-tuned $^{\textcolor{Green}{+}}_{\textcolor{red}{-}}$ & +TPS $^{\textcolor{Green}{+}}_{\textcolor{red}{-}}$ & +SSL $^{\textcolor{Green}{+}}_{\textcolor{red}{-}}$ \\
    \midrule
    \ChRep & 27.82M  & \textcolor{red}{4.0X} & 0.344  & - & - & -  \\
    \FxChn & \textcolor{red}{83.47M}  & 1.0X & 0.500 & 0.614 $_{\textcolor{Green}{22\%}}$  & 0.623 $_{\textcolor{Green}{1\%}}$ & 0.633 $_{\textcolor{Green}{2\%}}$ \\
    \midrule
    \Depth & 27.84M  & 1.0X & 0.517 & 0.616 $_{\textcolor{Green}{19\%}}$ & 0.622 $_{\textcolor{Green}{1\%}}$ & 0.608  $_{\textcolor{red}{2\%}}$ \\
    \SlPrm  & 27.86M & 1.0X  & 0.457 & 0.592 $_{\textcolor{Green}{29\%}}$ & 0.600 $_{\textcolor{Green}{1\%}}$ & 0.600 $_{\textcolor{Green}{0\%}}$ \\
    \TgPrm     & 27.84M & 1.0X  & 0.496 & 0.622 $_{\textcolor{Green}{25\%}}$ & 0.628 $_{\textcolor{Green}{1\%}}$ & 0.618 $_{\textcolor{red}{2\%}}$ \\
    \TmMix~\citep{savarese2018learning} & 28.02M  & 1.0X & 0.466 & 0.615 $_{\textcolor{Green}{31\%}}$ & 0.616  $_{\textcolor{Green}{0\%}}$ & 0.621 $_{\textcolor{Green}{1\%}}$  \\
    \HpNet~\citep{haHypernetworks2016} & 28.29M  & 1.0X  & 0.537 & 0.616 $_{\textcolor{Green}{14\%}}$  & 0.630 $_{\textcolor{Green}{2\%}}$ & 0.632 $_{\textcolor{Green}{0\%}}$ \\
    \bottomrule
  \end{tabular}
\end{table}

\subsection{Generalization capabilities}

One of the main goals of extending models to be channel-adaptive is that training with more data can generalize better to new domains or out-of-distribution data. There are two major generalization problems in microscopy: the first is biological and the second is technical. The biological generalization problem deals with experimental data that has novel biological conditions not seen during training. CHAMMI includes six biological generalization tasks with increasing difficulty, which introduce novel organelles, cell lines, and compounds in the prediction problems (Fig.~\ref{fig:tasks}). Generalization to new biological conditions (or domains) remains an open challenge of active research \citep{pernice2023out, wang2022generalizing,zhou2022domain}. The technical generalization problem deals with model performance on new datasets with different number of channels, which is a new challenge introduced with CHAMMI.

\textbf{Biological generalization}. We evaluate two training strategies that can be used to improve OOD generalization performance on the biological tasks. First, we consider Stochastic Weight Averaging Densely (\emph{SWAD})~\citep{cha2021swad}, which achieves generalization by seeking the flat minima of loss landscapes through dense weight averaging. 
Second, we consider Mutual Information Regularization with Oracle (\emph{MIRO})~\citep{cha2022domain}, which exploits the generalizability of a large-scale pre-trained model. 

\begin{figure}
  \centering
  \includegraphics[width=\linewidth]{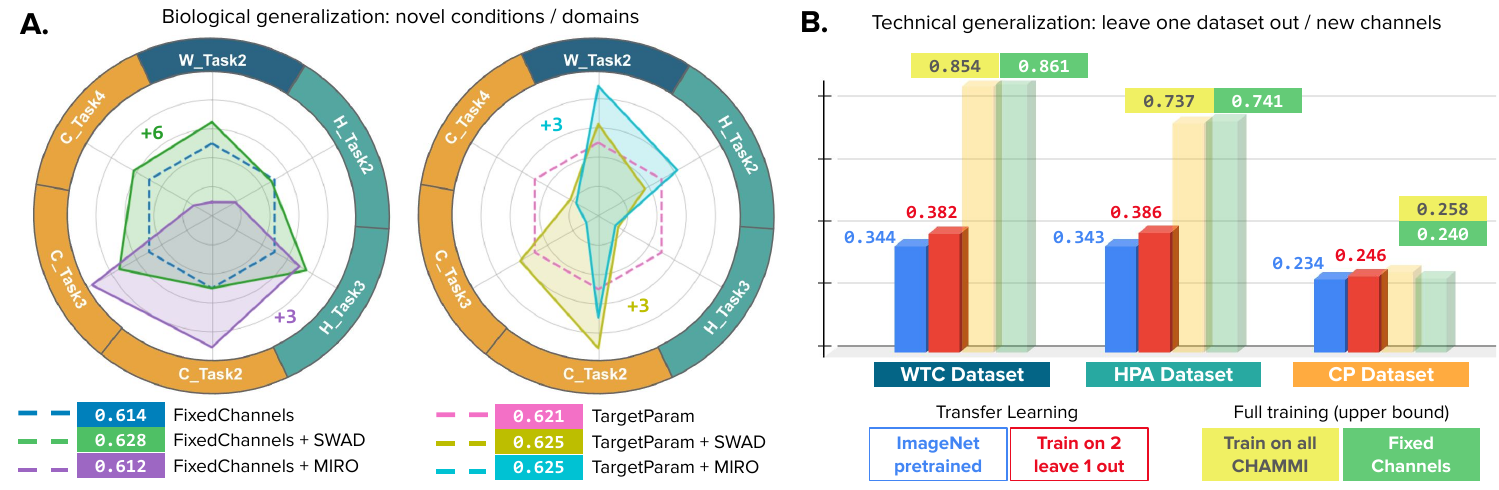}
  \caption{Generalization evaluation. A) Performance on the six biological generalization tasks of CHAMMI for two models: \FxChn~(left) and \TgPrm~(right). Each model is further trained with domain generalization methods SWAD~\cite{cha2021swad} and MIRO~\cite{cha2022domain}. B) Performance of training on two datasets and leaving one dataset out to evaluate technical generalization (avg. F1-score).}
  \label{fig:ood}
\end{figure}

First, we found that SWAD improves the performance of \FxChn\ on all biological generalization tasks, while MIRO improves performance on half of them (Fig.~\ref{fig:ood}A). The main weakness of MIRO is that it relies on a reference ImageNet pre-trained model as an oracle to learn image representations, which we have found to be suboptimal (Fig. \ref{fig:tasks}A). 
Next, we compare these results with \TgPrm\ trained in the standard way, and observe that it is able to perform similarly to \FxChn\ trained with SWAD or MIRO overall while improving in some tasks (appendix). This highlights the potential of channel-adaptive models to improve generalization performance out-of-the-box. Finally, we trained the \TgPrm\ model with SWAD and MIRO to evaluate the possibility of observing a synergistic effect of training with more data and optimizing the model with OOD techniques. \TgPrm\ slightly improves performance in three tasks out of six using SWAD or MIRO, which supports the idea that channel-adaptive models can improve generalization ability while also benefiting from advances in OOD generalization.

\textbf{Technical generalization.}
Here, we evaluate transfer learning performance of models trained with two datasets while testing on a holdout, third dataset that has a different number of channels not seen during training. The main hypothesis is that training a model with varying-channel images can perform better in a new image dataset with a different number of channels when no training data is available. From the five channel-adaptive models evaluated in this work, \Depth\ can be extended to a new set of channels without additional training by simply replicating / removing filters. Therefore, we evaluate the \Depth\ approach in this transfer learning setting, and use the same approach with ImageNet pre-trained weights as a baseline. Training on varying-channel images indeed improves performance up to 11\%, 12\%, and 5\% relatively, when the WTC, HPA, and CP datasets are held-out, respectively (Fig.~\ref{fig:ood}B). Note that the baseline model has only seen three-channel RGB images. For comparison, Fig.~\ref{fig:ood}B reports the performance of \Depth~and \FxChn~models trained with images from the evaluated dataset as an estimation of the upper-bound performance for transfer learning. The difference reveals ample room for improvement in methods and training strategies. All the results in this experiment are the average F1-score of the generalization tasks of each dataset (Sec.~\ref{sec:CHAMMI}).

\subsection{Limitations}
\textbf{Technical variation biases in the microscopy images.} Microscopy images, like other biological data, exhibit artifacts due to technical variation (\eg, equipment, microscope, technician, time of day). These artifacts can introduce positive biases that enhance performance in IID data and confounding factors that challenge accurate OOD generalization. The benchmark tasks were designed to include OOD data for improved estimation of model performance. However, technical variation may still play a role in confounding the results, which is also the topic of active concurrent research work.

\textbf{A relatively small-scale dataset compared to the actual needs of biological research.} Given the dataset balance, biological relevance, and known technical variations, our benchmark serves as a limited case study compared to the broader requirements of contemporary biological research. For instance, our tasks involve fewer than 10 classes, while real-world problems often encompass thousands of classes. We hope CHAMMI facilitates the rapid development and evaluation of models that can scale to larger varied-channel datasets that could be curated in the future, in which case the benchmark will still be relevant given its well-standardized set of tasks and metrics.
\section{Conclusion}

We presented CHAMMI, a new dataset for investigating a novel task: creating channel-adaptive models for flexible and scalable microscopy image analysis. We curated single-cell images from three publicly available sources, standardized their format, and designed nine biologically relevant downstream tasks to quantify the progress made by channel-adaptive models. We conducted extensive experiments that include baselines for channel-adaptive, non-adaptive, and domain generalization methods sampled from prior work on our task. We find that all channel-adaptive methods are able to improve or match the performance of non-adaptive specialists models at a smaller computational cost. Generalization tasks with OOD data benefit more from channel-adaptive models, and the \HpNet~model improved performance substantially with additional data augmentations and the incorporation of self-supervision. 

There are many open challenges to realize the full potential of channel-adaptive models in microscopy and beyond. First, exploring new architectures specifically designed for varying-channel images may further improve performance. Second, training and optimization strategies as well as data augmentations that target varying-channel images present many opportunities for innovation. Third, biological and technical generalization may benefit from novel domain adaptation techniques. Finally, creating unannotated, large-scale datasets (e.g., with many more microscopy modalities and diverse image resolutions) for pre-training channel-adaptive models can also complement our well-balanced and well-annotated benchmark. We leave these important research questions for future work, and expect the proposed CHAMMI benchmark to support many of these exciting developments. 

\textbf{Broader Impacts.} The acquisition and analysis of biological data have the potential for both positive and negative societal impacts when deployed in applications such as drug development and basic research. Bad actors may use this research to develop biotechnology that harms humans. We integrate openly accessible microscopy images acquired with different experimental conditions and technical formats. By solving the channel-adaptive model problem, this effort has the potential to bring together various microscopy imaging communities. Channel-adaptive models can also impact various other fields, including satellite, spectral, thermal, and ultrasound imaging, among others.

\begin{ack}
This study was supported, in part, by the Broad Institute Schmidt Fellowship program and by National Science Foundation NSF-DBI awards 2134695 and 2134696. Figure created with BioRender.com.
\end{ack}
{\small
\bibliographystyle{unsrtnat}
\bibliography{refs}
}
\newpage
\centerline{\huge{\textbf{Appendix}}}
\vskip 1cm

\textbf{Statement on ethical implications.} The CHAMMI dataset collects, processes, and shares open-access cellular microscopy images from publicly available sources. The CHAMMI benchmark provides a standardized API for loading and processing the dataset, as well as training and evaluating models on the benchmark. CHAMMI is under CC-BY 4.0 (Creative Commons Attribution 4.0 International Public) license. Please kindly ensure compliance with the Research Use Agreements when accessing either the curated dataset or the original datasets. Additionally, any bias inherent in the original datasets might be reflected in the curated CHAMMI dataset as well. The authors confirm that they bear all responsibility in case of violation of rights.

\section{CHAMMI: additional details}
\begin{figure}
  \centering
  \includegraphics[width=0.9\linewidth]{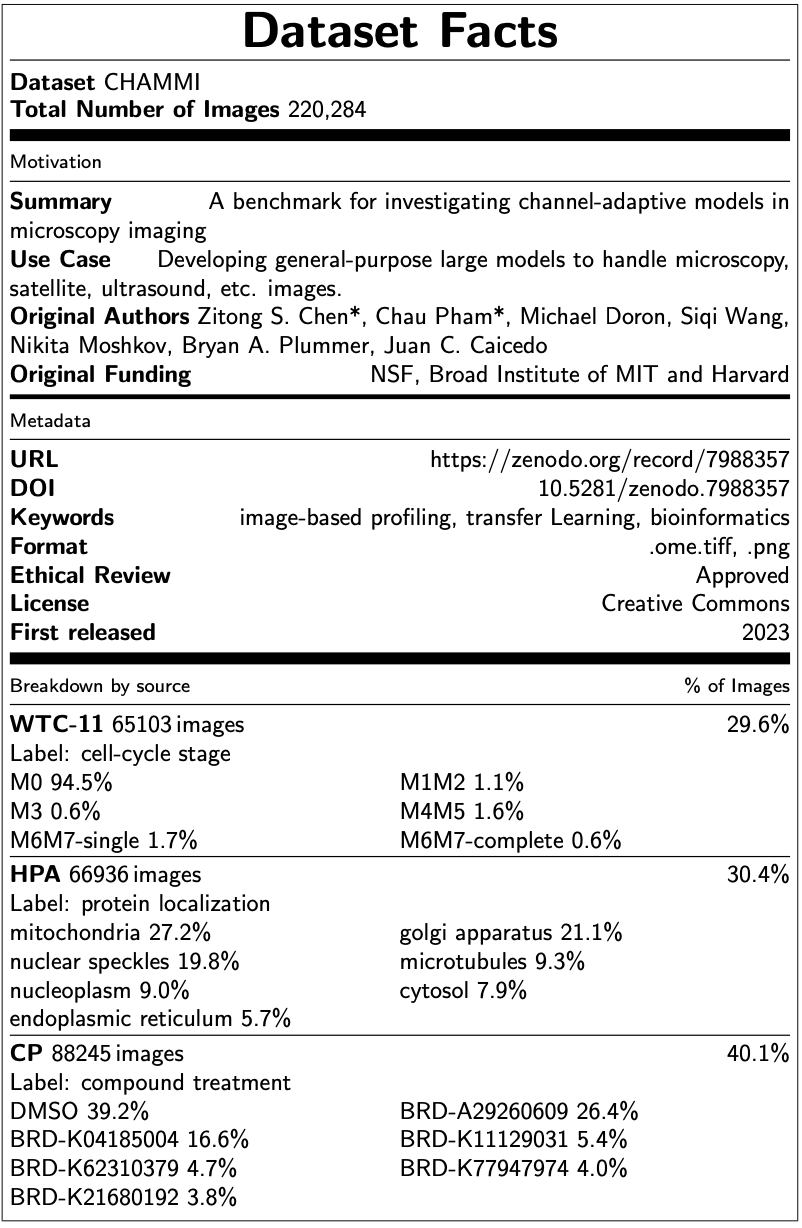}
  \vspace{-2mm}
  \caption{A dataset information card styled after nutrition labels, constructed based on ~\citep{nutrition_template, gebru2021datasheets}.}
  \label{fig:nutrition-label}
  \vspace{-2mm}
\end{figure}

The CHAMMI dataset contains single-cell resolution microscopy images curated from three main sources, with different sets of ground truth labels for each source dataset. A summary of the dataset is provided in Fig.~\ref{fig:nutrition-label}~\citep{nutrition_template, gebru2021datasheets}. 

\subsection{Data sources}
We describe the full license of the original datasets and the URL for  download.
\begin{enumerate}
    \item \textbf{WTC-11~\citep{viana2023integrated}} \\ 
    License: \href{https://www.allencell.org/terms-of-use.html}{Research Use Agree}\\
    URL: \href{https://open.quiltdata.com/b/allencell/packages/aics/hipsc_single_cell_image_dataset} {open.quiltdata.com/b/allencell/packages/aics/hipsc\_single\_cell\_image\_dataset}
    \item \textbf{HPA~\citep{thul2017subcellular}} \\ 
    License: CC-BY 3.0 \\
    URL: \href{https://www.kaggle.com/competitions/hpa-single-cell-image-classification/data}{www.kaggle.com/competitions/hpa-single-cell-image-classification/data}
    \item \textbf{Cell Painting~\citep{bray2017dataset,way2022morphology, gustafsdottir2013multiplex}} \\ 
    License: CC0 1.0 Universal \\
    URL-1: \href{https://bbbc.broadinstitute.org/BBBC022}{bbbc.broadinstitute.org/BBBC022} \\ 
    URL-2: \href{https://github.com/broadinstitute/cellpainting-gallery}{github.com/broadinstitute/cellpainting-gallery}

\end{enumerate}

\subsection{Data processing}
\textbf{WTC-11 dataset.} The original dataset contains 214,037 human induced pluripotent stem cells from 25 isogenic cell lines. Each cell line contains fluorescent tagging for one protein via CRISPR/Cas9 gene editing to reference a cellular compartment. We selected 65,103 cells by filtering for cells with fluorescent-protein (FP) taggings for one of seven cellular compartments: nuclear speckles, mitochondria, microtubules, Golgi apparatus, nucleoplasm, cytosol, and endoplasmic reticulum (ER). We used the max z-channel projection of the three fluorescent channels to project the 3D images into 2D planes. Cell segmentation was performed using the masks provided by the original authors \citep{viana2023integrated}, which were obtained with the Allen Cell and Strucutre Segmenter~\citep{chen2018allen}. We then normalize the images into between 0 and 255 pixel intensity and unfolded the three channels by concatenation. These result in 65,103 single-cell, single-channel images of size 1122 x 238.

\textbf{HPA.} The original dataset is used in the Kaggle competition ‘Human Protein Atlas - Single Cell Classification’~\citep{hpa-single-cell-image-classification} and is composed of field-of-view images with cells from 28 cell lines. The cells were treated with fluorescent dyes that bind to the nucleus, microtubules, endoplasmic reticulum (ER), and a protein of interest, resulting in four-channel images. Cell segmentation is performed using the HPA-Cell-Segmentation algorithm~\citep{hpacellseg} recommended in the Kaggle challenge. We take the bounding box with the mask and crop the images to 512x512 around the center of the cell. We then normalized the images to 0 and 255 pixel intensity. We also filtered for cells with subcellular protein localizations in one of seven cellular compartments: nuclear speckles, mitochondria, microtubules, Golgi apparatus, nucleoplasm, cytosol, and ER. Finally, we unfolded the four-channel images by concatenation. These procedures result in 66,936 single-cell images of size 2048x512.

\textbf{Cell Painting.} The original dataset consists of three sources: BBBC022~\citep{gustafsdottir2013multiplex}, CDRP~\citep{bray2017dataset}, and LINCS~\citep{way2022morphology}. Images are acquired using the Cell Painting protocol, with six fluorescent dyes staining eight cellular compartments. BBBC022 and CDRP are both compound-screening datasets tested on U2OS cells, including 1600 and 30616 single-dose compounds respectively. LINCS is a compound screen of 1249 drugs across six doses on A549 cells. Among the common compounds between the three datasets, we selected six single-dose compounds and DMSO negative control to include in CHAMMI. Of the six compounds, two of each are shown to have weak, median, and strong morphological effects on cell morphology compared to the negative control (Tab.~\ref{tab:compound}). We ensured that all six compounds have different mechanisms of action and that the selected concentrations across the three datasets have minimal differences. CellProfiler~\citep{carpenter2006cellprofiler} is used to segment the cells, with global Otsu thresholding in the nucleus channel, followed by cell body segmentation with the watershed method in the ER/RNA channel. We cropped the images to size 160x160 centered on the nucleus of each cell without masking so as to preserve the context of single cells. We then unfolded the five channels to get $88,245$ images of size $800 \times 160$.

\setlength\tabcolsep{2pt}
\begin{table*}
  \caption{Information about selected compound treatments for Cell Painting images. Effect: the ability of the compound to induce morphological changes. Strong means that the compound induces significant morphological changes, and \textit{vice versa}. Concentration: the selected concentration of the compound in each source dataset. MOA: ground truth annotation~\citep{repurposing} of mechanism of action. Present in: the training and/or testing sets the label is present in.} 
  \label{tab:compound}
  \centering
  \scriptsize\begin{tabular}{*{8}{c}}
    \toprule
    & \multicolumn{3}{c}{ }
    & \multicolumn{1}{c}{Concentration} \\
    & \multicolumn{3}{c}{ }
    & \multicolumn{1}{c}{(mmol/L))} \\
    \cmidrule(r){4-6}
    Broad ID & Name & Effect & BBBC022 & CDRP & LINCS & MOA & Present in\\
    \midrule
    BRD-A29260609 & acebutolol & weak & 2.68 & 5.00 & 3.33 & adrenergic receptor antagonist & Train, Task 1, 2, 3 \\
    BRD-K11129031 & gemfibrozil & weak & 3.99 & 5.00 & 3.33 & lipoprotein lipase activator & Task 4 \\
    BRD-K04185004 & oxybuprocaine & medium & 2.90 & 2.90 & 3.33 & local anesthetic & Train, Task 1, 2, 3 \\
    BRD-K62310379 & fluticasone-propionate & medium & 2.00 & 2.00 & 1.11 & glucocorticoid receptor agonist & Task 4 \\
    BRD-K21680192 & mitoxantrone & strong & 1.93 & 1.93 & 1.11 & topoisomerase inhibitor & Train, Task 1, 2, 3 \\
    BRD-K77947974 & fluspirilene & strong & 5.26 & 5.26 & 3.33 & dopamine receptor antagonist & Task 4 \\
    \bottomrule
  \end{tabular}
  \vspace{-2mm}
\end{table*}

\subsection{Data sampling and splitting}
Due to the biological nature of microscopic images, many of the standard classification tasks involve using inherently imbalanced datasets. This is the case for the three datasets we have chosen as well. While it is possible to balance the data by upsampling or downsampling images, we decided to preserve the original distribution of classes so as to simulate a real-life biological application setting. Meanwhile, we keep the ratio of classes and other biological annotations consistent across training and testing. Results confirm that supervised models were able to learn both the majority and minority classes with high accuracy under this configuration. Additionally, we ensure that the ratio between the size of each testing set and the combined training set has a ratio between 1:9 and 1:4. We choose to be more flexible about the training over testing ratios due to the small sample size of certain classes and the comparatively much larger sample size of others.

\textbf{WTC-11.} This subset of CHAMMI uses cell cycle stages as the classification label, which results in an inherent imbalance of data. Natural cells spend approximately 90\% of their time in interphase, which means that the majority of cells captured in an image will be in interphase. We preserve the original distribution of labels in our dataset by random sampling images within classes into training and testing to ensure that the ratio between all the classes is consistent. Additionally, since the protein channel of each image contains FP taggings for one of seven cellular compartments (see Section A.2), we also used stratified random sampling to ensure a consistent ratio of FP-tagged cellular compartments across training and testing sets.

\textbf{HPA.} The HPA images in CHAMMI use subcellular protein localization as the classification label, which is also inherently imbalanced. Since proteins can localize at multiple locations, and a few compartments host much more proteins than others, the original HPA images are multi-labeled and have inherent label imbalance. We first filtered for images with only one cellular localization annotation (to ensure single-labeled data) that is in the list of the seven chosen cellular compartments. After filtering, the remaining images come from 18 different cell lines. We then stratified them into training and testing sets to keep the label and cell line ratio consistent across sets. 

\textbf{CP.} The Cell Painting images in CHAMMI use compound treatment as the classification label, which is also imbalanced. Since we selected two strong, median, and weak effect compounds, the number of viable cells surviving the strong treatment is bound to be lower than the number of cells surviving the weak treatment. Additionally, Cell Painting images are known to be affected by the batch effect of plates and datasets, which means images taken from different plates can look very different despite having similar cell morphology. Therefore, we used stratified sampling to ensure that the ratio of labels, source dataset, and plates is consistent across training and testing sets.  

\section{Experiments}
In this section, we present the experimental details including model architecture, hyperparameter tuning, training schedule, etc. The code we used for training and evaluating models is publicly available at 
\url{https://github.com/chaudatascience/channel_adaptive_models}
\subsection{Model architecture}
\begin{figure}
  \centering
  \includegraphics[width=\linewidth]{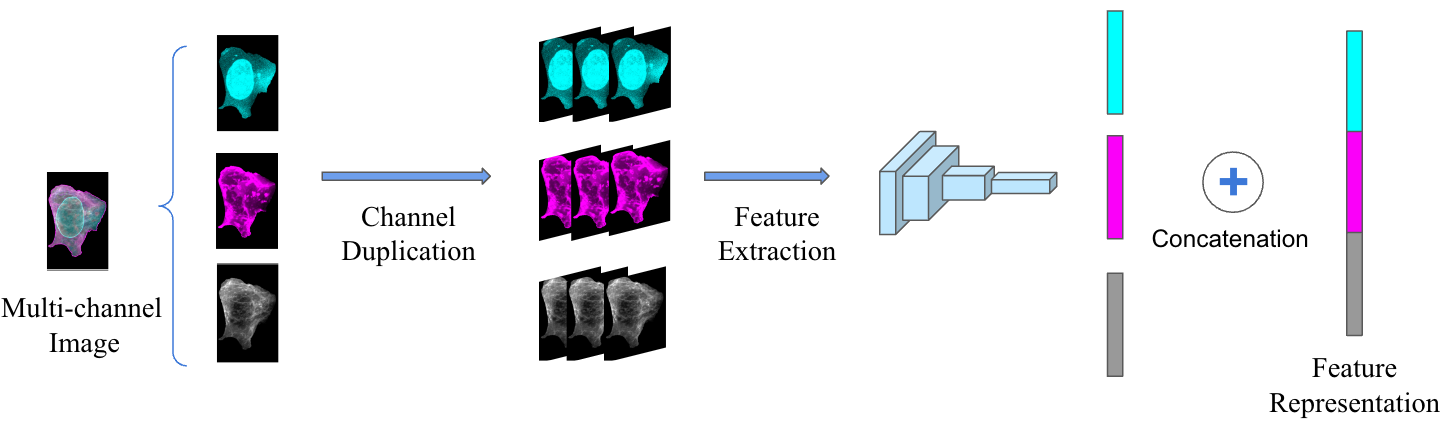}
  \vspace{-4mm}
  \caption{\textbf{ChannelReplication model architecture}. For a multi-channel image (\eg, \textit{WTC} in this example), each channel is replicated to form a 3-channel image as the input for the ConvNeXt~\cite{liu2022convnet} model. The final representation of the multi-channel image is the concatenation of the feature output of each channel.
}
  \label{fig:channel_dup_model}
  \vspace{-4mm}
\end{figure}

\begin{figure}
  \centering
  \includegraphics[width=\linewidth]{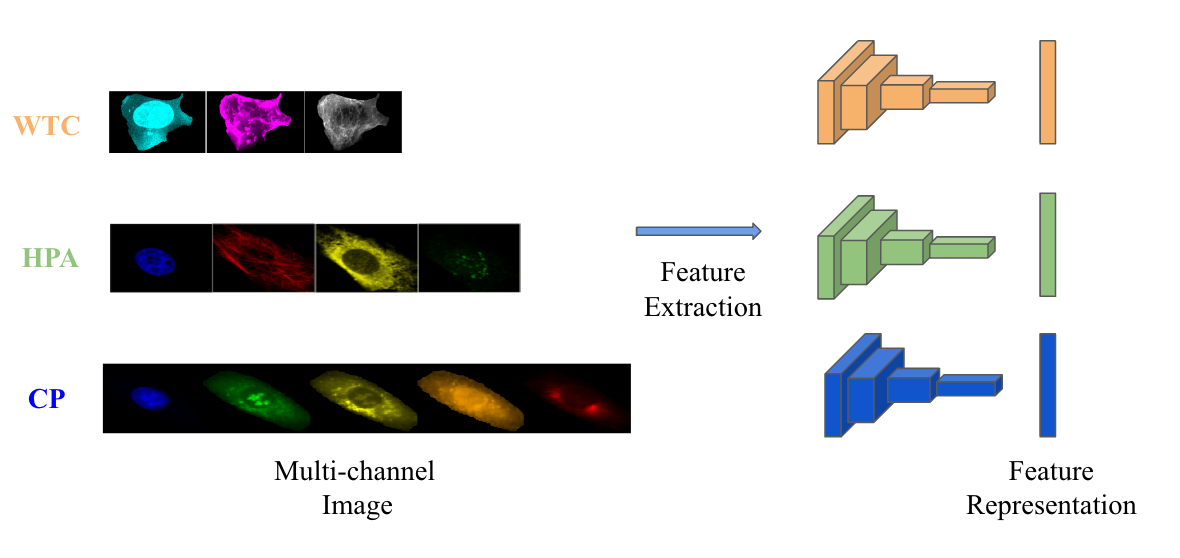}
  \vspace{-2mm}
  \caption{\textbf{FixedChannels model architecture}. Each dataset has its own single network. Note that the first convolutional layers of these networks are adjusted based on the specific number of input channels presenting in each dataset (\ie, duplicating channel weights if needed).
}
  \label{fig:fix_channel_model}
  \vspace{-4mm}
\end{figure}

\begin{figure}
  \centering
  \includegraphics[width=\linewidth]{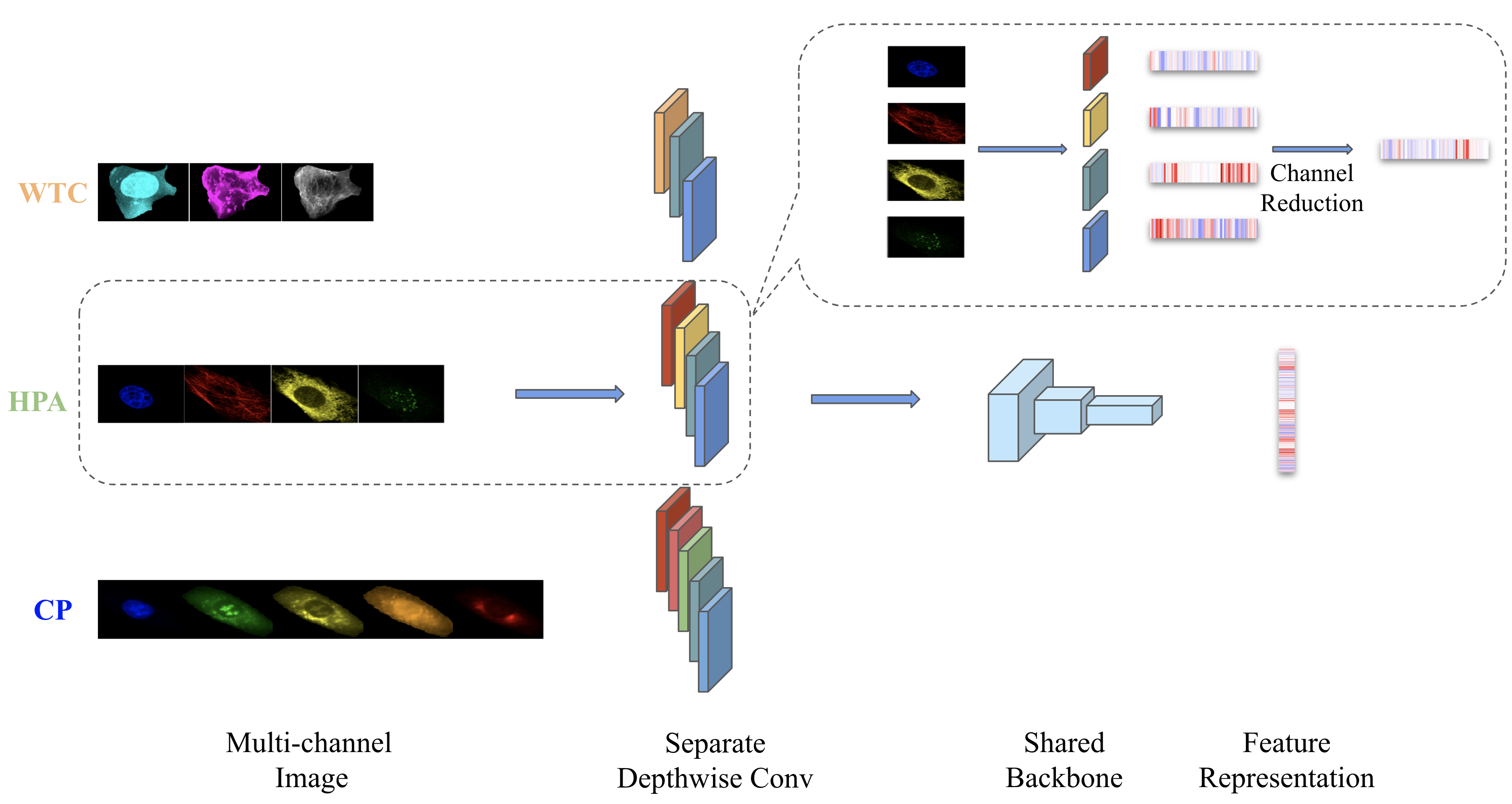}
  \vspace{-4mm}
  \caption{\textbf{Depthwise model architecture.} A single convolutional filter is applied for each input channel. The feature outputs from all channels are reduced to a single feature (channel reduction) as the input to the shared backbone network. This enables the shared backbone to receive a fixed size of input regardless of the number of input channels presenting in each dataset.
}
  \label{fig:depthwise_model}
  \vspace{-2mm}
\end{figure}

\begin{figure}
  \centering
  \includegraphics[width=\linewidth]{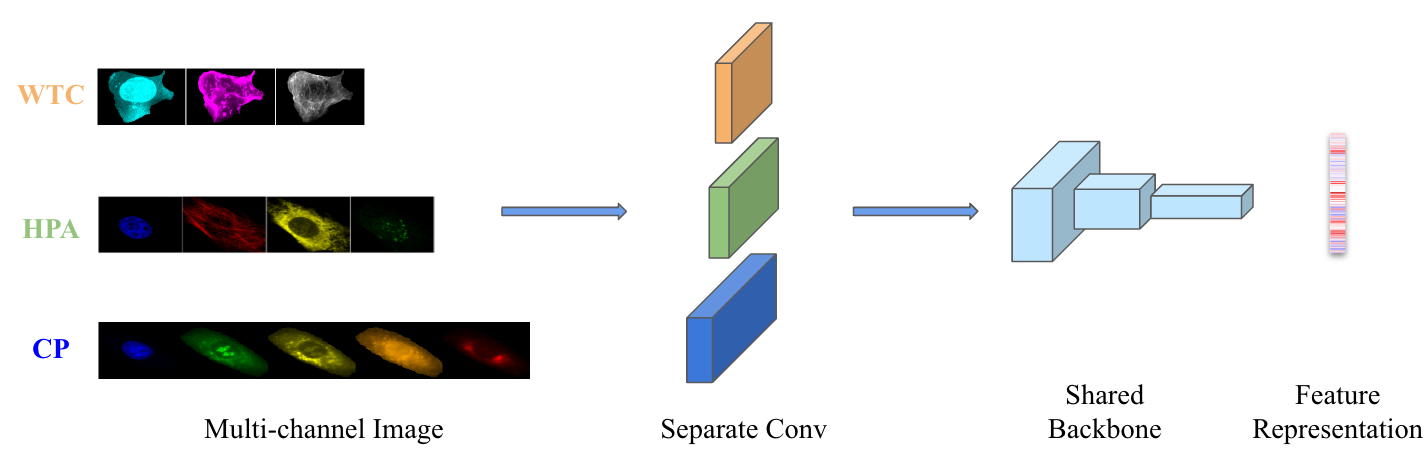}
  \vspace{-4mm}
  \caption{\textbf{TargetParam model architecture.} The network consists of individual convolutional heads, each dedicated to a particular sub-dataset, along with a shared backbone.  When processing a multi-channel image, the image is assigned to a specific convolutional head based on its number of channels. Although these heads receive images with varying number channels as input, they produce features of the same size.  These features are then fed through the shared backbone to obtain the final representation.
}
  \label{fig:shared_model}
  \vspace{-2mm}
\end{figure}

\begin{figure}
  \centering
  \includegraphics[width=\linewidth]{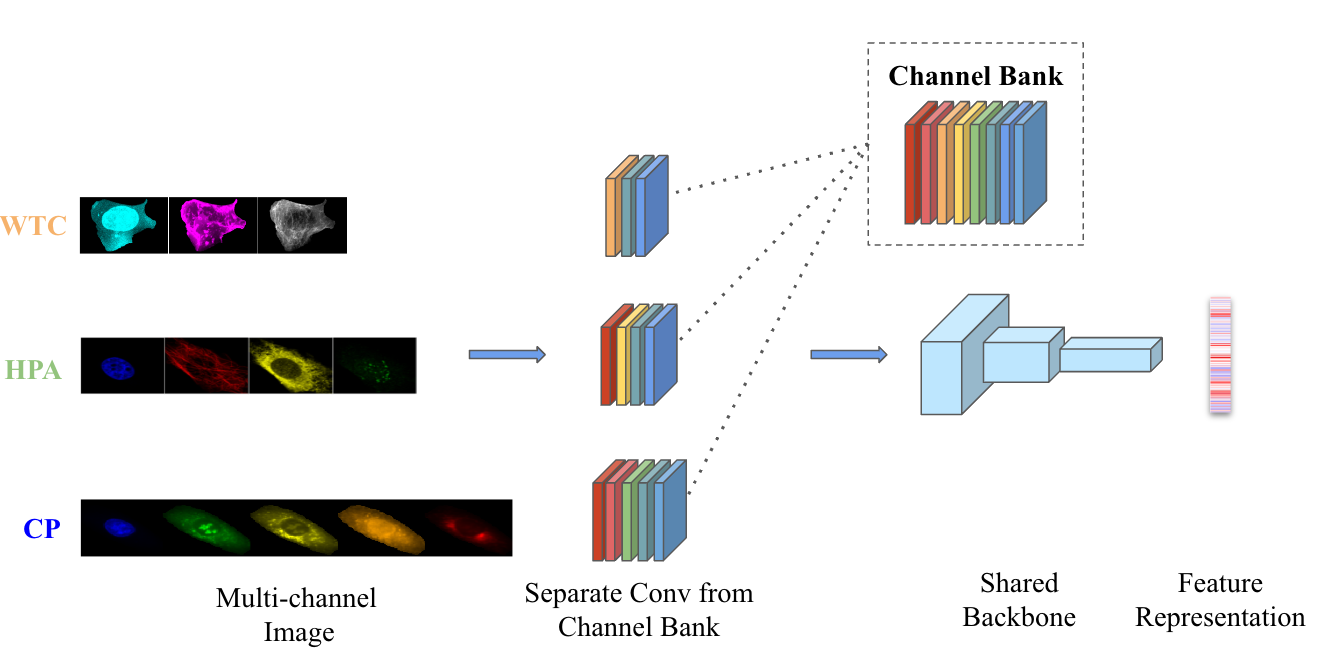}
  \vspace{-4mm}
  \caption{\textbf{SliceParam model architecture}. The network consists of a channel bank and a shared backbone.  Each distinct channel within the dataset has its own filter within the channel bank. When processing a multi-channel image, it is assigned to a specific convolutional layer that combines the corresponding channel filters from the channel bank. The resulting output features whose sizes are fixed, are then passed through the shared backbone to get the final representation.
}
  \label{fig:slice_model}
  \vspace{-2mm}
\end{figure}

\begin{figure}
  \centering
  \includegraphics[width=\linewidth]{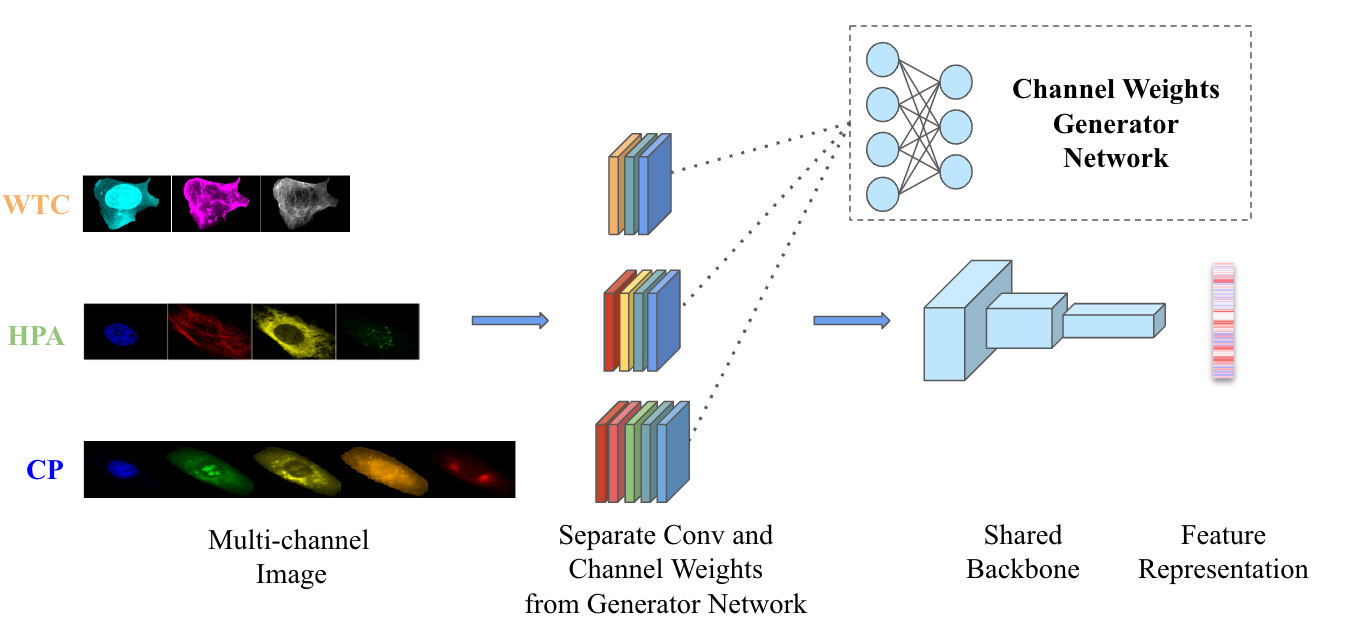}
  \vspace{-4mm}
  \caption{\textbf{HyperNet~\citep{haHypernetworks2016} model architecture.} The network comprises a channel weights generator network and a shared backbone. Each distinct channel within the dataset is represented by a trainable channel embedding in the generator network. These embeddings are utilized to generate the weights for the corresponding input channel. The weights for different channels are concatenated to form a comprehensive set of weights used to process the image. The resulting features are then passed through the shared backbone network, ultimately producing the final representation.
}
  \label{fig:hypernet_model}
  \vspace{-2mm}
\end{figure}

\begin{figure}
  \centering
  \includegraphics[width=\linewidth]{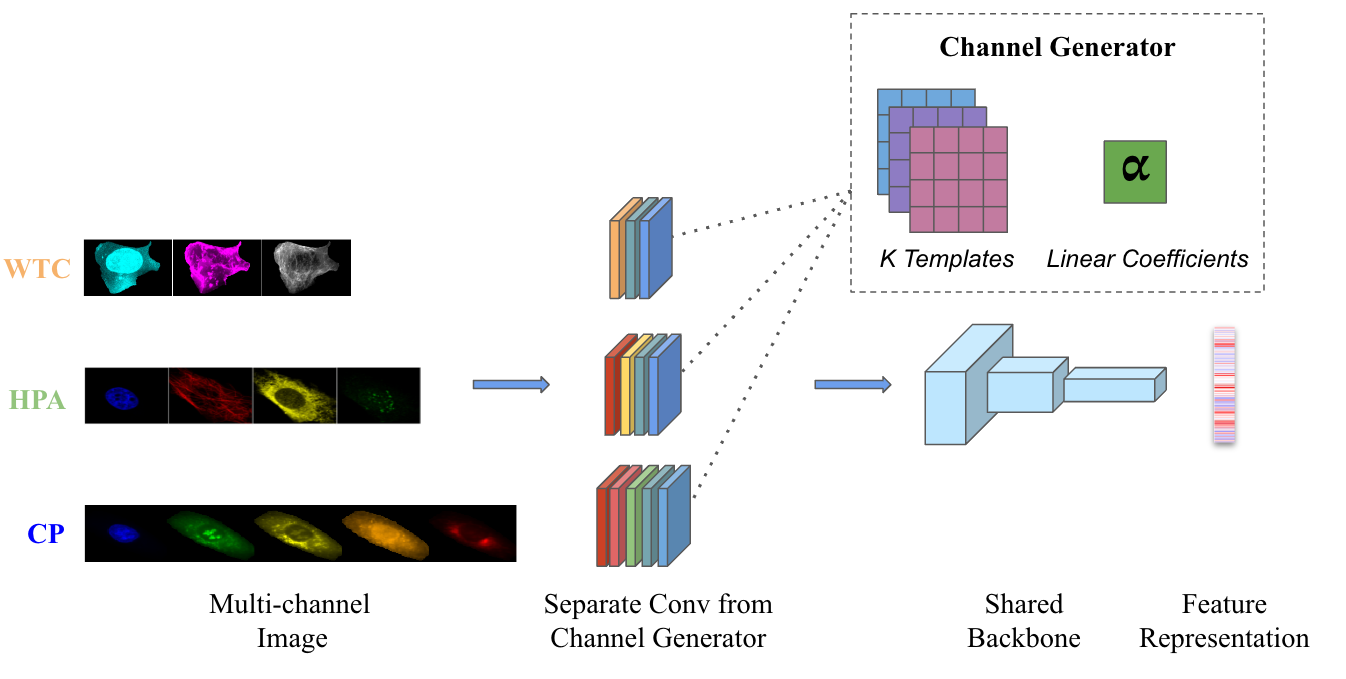}
  \vspace{-4mm}
  \caption{\textbf{Template mixing~\citep{savarese2018learning,plummerNPAS2022} model architecture.} The network comprises a channel generator with a set of templates, linear coefficients, and a shared backbone. Each distinct channel in the dataset has a linear coefficient, which is a $k$-dimensional vector, where $k$ is the number of templates. The convolutional filter for each distinct channel is obtained by a linear combination of the templates using its corresponding linear coefficient. As a result, a multi-channel image is assigned to a specific convolutional layer generated by the channel generator. The resulting features are then fed into a shared backbone, resulting in the final representation.
}
  \label{fig:template_model}
  \vspace{-2mm}
\end{figure}

In all experiments, we used a ConvNeXt~\cite{liu2022convnet} model pre-trained on ImageNet 22K~\cite{deng2009imagenet} as the backbone. We considered three extensions of the convolutional network and adapted two previously proposed strategies as potential solutions to the channel-adaptive problem. See Fig.~\ref{fig:channel_dup_model}-Fig.~\ref{fig:template_model} for an illustration of the different architectures.

\subsection{Representation learning}
In this experiment, we employ ConvNeXt~\cite{liu2022convnet} as the underlining backbone for our baseline models. While the initial layers may differ across models, all of them share the same backbone architecture. To extract the image representations, we remove the classifier head and only keep the feature extractor. 

It is worth noting that the embedding vectors obtained from the backbone have a dimension of $7 \times 7 \times 768$, which makes computing Euclidean distances computationally expensive. To address this issue, we employ pooling techniques to reduce the dimensionality to a more manageable 768-dimensional space. Although we use adaptive average pooling as our default method, we have observed minimal differences when switching to other pooling methods such as adaptive max pooling or combining both.

\subsection{Loss functions}
During the inference, our goal is to use the trained model to acquire a meaningful representation for each input image, and the model can be used for both familiar and novel classes. To achieve this, we employ the ProxyNCA++ loss, introduced by Teh et al.~\citep{teh2020proxynca}, which is formulated as follows:
$$
L_{\text {ProxyNCA++ }}=-\log \left(\frac{\exp \left(-d\left(\frac{x_i}{\left\|x_i\right\|_2}, \frac{f\left(x_i\right)}{\left\|f\left(x_i\right)\right\|_2}\right) * \frac{1}{T}\right)}{\sum_{f(a) \in A} \exp \left(-d\left(\frac{x_i}{\left\|x_i\right\|_2}, \frac{f(a)}{\|f(a)\|_2}\right) * \frac{1}{T}\right)}\right)
$$

Where $d(x_i, x_k)$ is Euclidean squared distance computed on feature embedding, $A$ denotes the set of all proxies, and $T$ is temperature. During the training process, each training class is represented as a proxy. We compare input images to the proxies, with an objective to draw samples toward their corresponding proxies while simultaneously pushing them away from all other proxies.

When the value of temperature $T$ is equal to 1, we obtain a standard Softmax function. As $T$ decreases, it results in a more concentrated and peaky probability distribution. We perform fine-tuning of the temperature by utilizing a low temperature setting, specifically ranging from 0.05 to 0.5.

To address the out-of-distribution (OOD) issues, we have also considered self-supervised approaches~\cite{chen2020simple,chen2020big,he2020momentum, grill2020bootstrap, jiang2023hierarchical,caron2021emerging,oquab2023dinov2}. In our training process, each image undergoes two transformations, each involving different operations such as random cropping, flipping, and TPS~\cite{tang2019augmentation}. To incorporate self-supervised learning, we adapted the SimCLR~\cite{chen2020simple,chen2020big} framework and introduced an extra loss term that enforces similarity between positive pairs, where positive pairs are represented by different augmentations of the same image. The self-supervised learning (SSL) loss is formulated as follows:
$$
L_{\text{SSL}}=-\log \frac{\exp sim(x_i, x_j)/T}{\sum_{k=1}^{2N} \exp sim(x_i, x_k)/T}
$$

where $N$ is the number of examples within a batch, $sim(., .)$ denotes the cosine similarity between two image representations. In addition, $T$ is the temperature parameter, and $\{x_i$, $x_j\}$ is a positive pair (i.e., augmented from the same image). The final SSL loss is computed over all the positive pairs in the training set. Note that since every image has two augmentations, the total number of examples is $2N$.

The combined loss is the combination of proxy loss and SSL loss, where $\alpha \in [0,1]$. We reported results from experiments where $\alpha=0.2$.
$$
L_{\text{combined}} = \alpha \times L_{\text{SSL}} + (1- \alpha) \times L_{\text {ProxyNCA++ }}
$$

\subsection{Implementation details}
In our experiments, we utilize the ConvNeXt~\cite{liu2022convnet} tiny version, which is pretrained on the ImageNet 22k dataset~\cite{deng2009imagenet}. We adapt the implementation from the repository provided by Hugging Face\footnote{https://github.com/huggingface/pytorch-image-models}. To fine-tune the model, we run 15 epochs on a single GPU using the AdamW optimizer~\cite{loshchilov2017decoupled} with a momentum of 0.9. The betas for AdamW are set to 0.9 and 0.999, and we apply a weight decay of $5 \times 10^{-4}$. Our batch size is set to 128. Note that when training on CHAMMI, each batch consists of images from all three datasets. This mixing training ensures that the model can effectively leverage the shared information presenting across the datasets throughout the training phase.

To determine the optimal learning rate, we sweep over some values in a range of values from $1.0 \times 10^{-6}$ to $1.0 \times 10^{-3}$ on a logarithmic scale. We use a cosine schedule to gradually reduce the learning rate to $1.0 \times 10^{-7}$ at the end of the training process. As our objective, we utilize the ProxyNCA++ loss~\citep{teh2020proxynca} function.

MIRO~\cite{cha2022domain}: To compute the  regularization term in MIRO, we extract intermediate outputs by each model block, i.e., stem output, stage 1, 2, and 3 from ConvNeXt~\cite{liu2022convnet} backbone. The final loss is a linear combination of ProxyNCA++ loss~\citep{teh2020proxynca} and the regularization term scaled by coefficient weight $\lambda$. We incorporate the implementation of the authors~\footnote{https://github.com/kakaobrain/miro} into our codebase.

SWAD~\cite{cha2021swad}: We utilize the implementation of SWAD~\footnote{pytorch.org/blog/pytorch-1.6-now-includes-stochastic-weight-averaging} from the Pytorch library.

\textbf{Compute resources}: For this study,  each experiment was run on a single NVIDIA RTX A6000 (48GB RAM) and three Intel(R) Xeon(R) Gold 6226R CPU @ 2.90GHz. 

\subsection{Evaluation}
In the testing stage, we use a one nearest-neighbor (1-NN) algorithm with cosine similarity as the distance metric to predict the label for each test sample. The computations were performed using Faiss~\footnote{https://github.com/facebookresearch/faiss}, a software framework designed to facilitate efficient searching for similarities and clustering of dense vectors. 

For tasks H\_Task3 and C\_Task4, test images do not share labels with the training images, so we apply a leave-one-out strategy. Specifically, we organize the test data into sub-groups based on non-label annotations (\textit{e.g.} cell type, source dataset). Each sub-group includes samples from all label classes in the testing set. During the evaluation, we hold out one sub-group for prediction and compute 1-NN search on both training images and non-holdout testing images. Note that this strategy still keeps the test data out of training deep learning models to prevent data leakage. H\_Task3 uses cell type for sub-group division in the leave-out procedure since there are 17 cell lines in this test set; C\_Task4 uses plate ID since cells in this test set come from four different plates. Since the models were not trained on images with the novel test labels, leave-out tasks are significantly harder than the other standard nearest-neighbor tasks due to distribution shifts.

\section{Additional results and discussions}
\label{sectionC-task-level}

\textbf{Task-level results.} 
We provide the F1 scores on each task for trained and fine-tuned models in Tab.~\ref{tab:full-f1} and report the scores in the \enquote{Average OOD - Mean} column in Fig.4 of the main paper. Models with the highest score for each column are printed in bold. We observe that all models benefit from ImageNet pre-training. Channel adaptive models, which are computationally less costly during both training and testing (Tab.~1 in the main paper), show comparable or superior performances as the baseline models, with the fine-tuned TargetParam model achieving the highest score on average.

\begin{table*}[t]
  \caption{\small Task-wise F1 scores on the validation set of CHAMMI. Average scores are computed by taking the mean of F1 score within each source dataset (see Section \ref{sectionC-task-level}), and then averaging across datasets. OOD: out-of-distribution (\textit{i.e}. generalization tasks).}
  \label{tab:full-f1}
  \centering
  \scriptsize\begin{tabular}{r*{13}{c}}
    \toprule
    & \multicolumn{4}{c}{Average OOD} 
    & \multicolumn{2}{c}{WTC} 
    & \multicolumn{3}{c}{HPA} 
    & \multicolumn{4}{c}{CP}   \\
    \cmidrule(r){2-5} \cmidrule(r){6-7} \cmidrule(r){8-10} \cmidrule(r){11-14}
    Model & Mean & WTC & HPA & CP & Task1 & Task2 & Task1 & Task2 & Task3 & Task1 & Task2 & Task3 & Task4 \\
    \midrule
    ChannelReplication & 0.344	& 0.422	& 0.336	& 0.275	& 0.594	& 0.422	& 0.560	& 0.418	& 0.254	& 0.839	& 0.478	& 0.223	& \textbf{0.122}    \\
    FixedChannels - Trained & 0.500	& 0.648	& 0.592	& 0.259	& 0.649	& 0.648	& 0.807	& 0.763	& 0.421	& 0.660	& 0.481	& 0.230	& 0.066      \\
    FixedChannels - Fine-tuned &  0.614& \textbf{0.861} &	0.741&	0.240 & \textbf{0.881}	&\textbf{0.861}	&\textbf{0.950}	& \textbf{0.932}	&0.551	&0.941	&0.484	&0.123	&0.112\\
    \midrule
    Depthwise - Trained       & 0.517 & 0.652 & 0.644 & 0.256 & 0.689	& 0.652	& 0.849	& 0.813	& 0.475	& 0.673	& 0.478	& 0.224	& 0.065 \\
    Depthwise - Fine-tuned    & 0.616 &	0.854 & 0.737 & 0.258 & 0.880	& 0.854	& 0.934	& 0.918	& 0.557	& 0.927	& 0.520	& 0.172	& 0.081\\
    TargetParam - Trained     & 0.496 & 0.590	& 0.623	& 0.273	& 0.695	& 0.590	& 0.837	& 0.794	& 0.452	& 0.717	& 0.508	& 0.234	& 0.077 \\
    TargetParam - Fine-tuned  & \textbf{0.622} & 0.843 & \textbf{0.760} & 0.264 & 0.879	& 0.843	& 0.945	& 0.925	& \textbf{0.594}	& \textbf{0.946}	& 0.512	& 0.174	& 0.107\\
    SliceParam - Trained      & 0.457 & 0.568	& 0.546	& 0.256	& 0.616	& 0.568	& 0.770	& 0.690	& 0.403	& 0.646	& 0.475	& 0.222	& 0.071 \\
    SliceParam - Fine-tuned   & 0.592 &	0.751 &	0.742 & \textbf{0.282}	&0.844	& 0.751	& 0.927	& 0.902	& 0.582	& 0.902	& \textbf{0.573}	& 0.202	& 0.072\\
    HyperNet~\citep{haHypernetworks2016} - Trained      & 0.537	& 0.661	& 0.671	& 0.278	& 0.726	& 0.661	& 0.887	& 0.858	& 0.483	& 0.720	& 0.517	& \textbf{0.247} & 0.069 \\
    HyperNet~\citep{haHypernetworks2016} - Fine-tuned  & 0.616 &	0.846	& 0.745	& 0.257	& 0.864	& 0.832	& 0.941	& 0.912	& 0.584	& 0.923	& 0.542	& 0.180	& 0.077 \\
    Template mixing~\citep{savarese2018learning} - Trained    & 0.466 & 0.565 & 0.577	& 0.257	& 0.631	& 0.565	& 0.808	& 0.741	& 0.413	& 0.671	& 0.468	& 0.227	& 0.075 \\
    Template mixing~\citep{savarese2018learning} - Fine-Tuned & 0.615 & 0.823 &	0.743	 & 0.279 & 0.855	& 0.823	& 0.939	& 0.919	& 0.566	& 0.906	& 0.542	& 0.202	& 0.094 \\
    \bottomrule
  \end{tabular}
  \vspace{-2mm}
\end{table*}

\begin{figure}
  \centering
  \includegraphics[width=\linewidth]{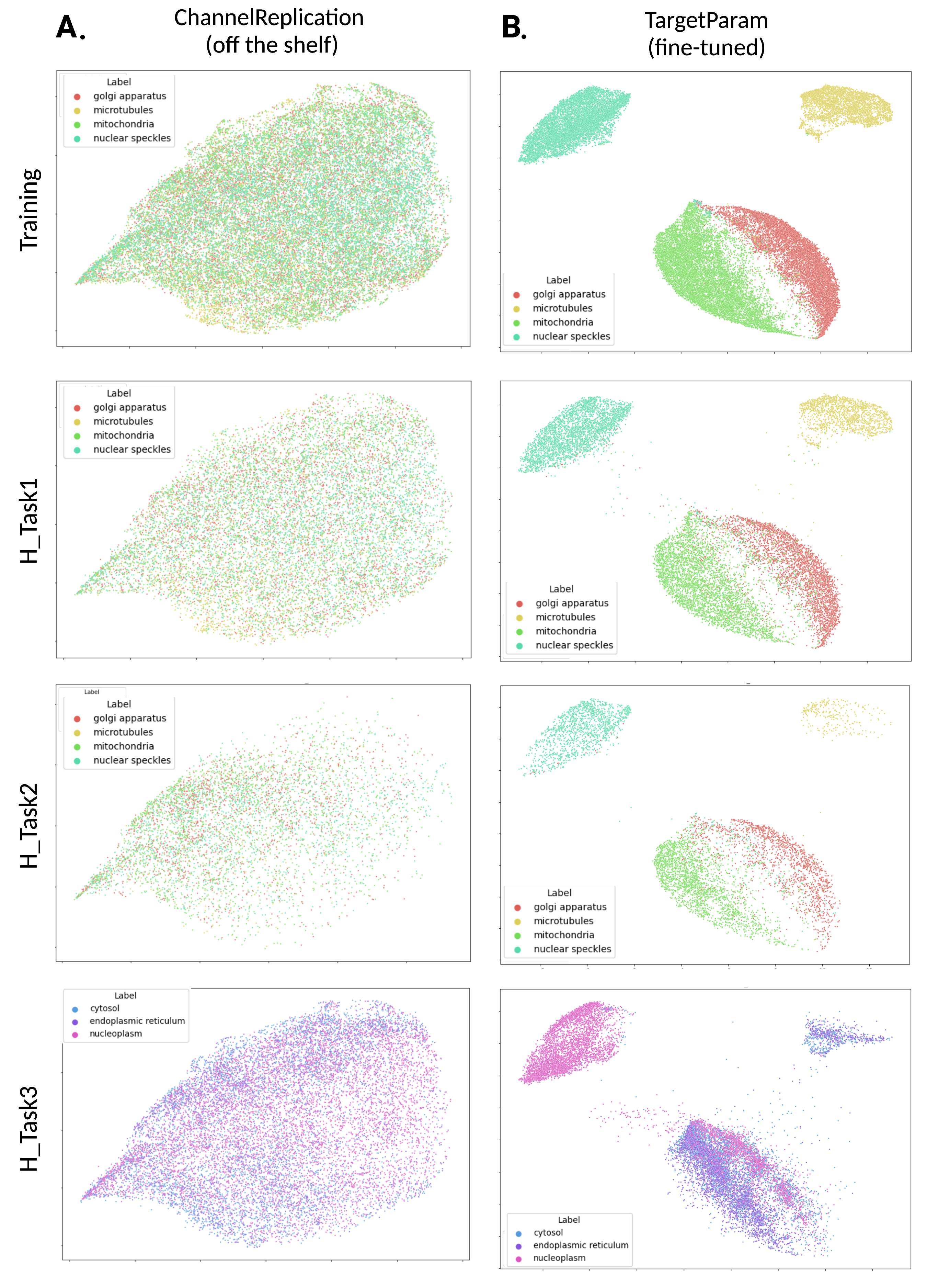}
  \vspace{-2mm}
  \caption{UMAP visualization of ChannelReplication and TargetParam features from the HPA subset of CHAMMI. Each point represents a single cell and colors represent the classification labels (protein localization). We compare the features extracted with off-the-shelf ChannelReplication model (A) pre-trained on ImageNet 22K~\cite{deng2009imagenet} and with fine-tuned TargetParam model (B) after training on CHAMMI.
}
  \label{fig:umap}
  \vspace{-2mm}
\end{figure}

\textbf{UMAP representation } We evaluated the clustering of features extracted by each model with UMAP visualizations and included the results for the TargetParam model on HPA data in Fig.~\ref{fig:umap}. We trained the UMAP with the CHAMMI HPA training set and projected the test sets without training. In Fig.~\ref{fig:umap}A, features are extracted with an off-the-shelf ChannelReplication model evaluated directly after pre-training on ImageNet 22K without fine-tuning on CHAMMI. In Fig.~\ref{fig:umap}B, the TargetParam model is pre-trained on ImageNet and fine-tuned with CHAMMI. Each row represents cells from a training or testing set of CHAMMI HPA, colored by their classification label (\ie, protein localization). We observe clustering based on protein localization in the test sets in the fine-tuned TargetParam model but not in the off-the-shelf model, which highlights the necessity of fine-tuning.

\begin{figure}[t]
  \centering
  \includegraphics[width=\linewidth]{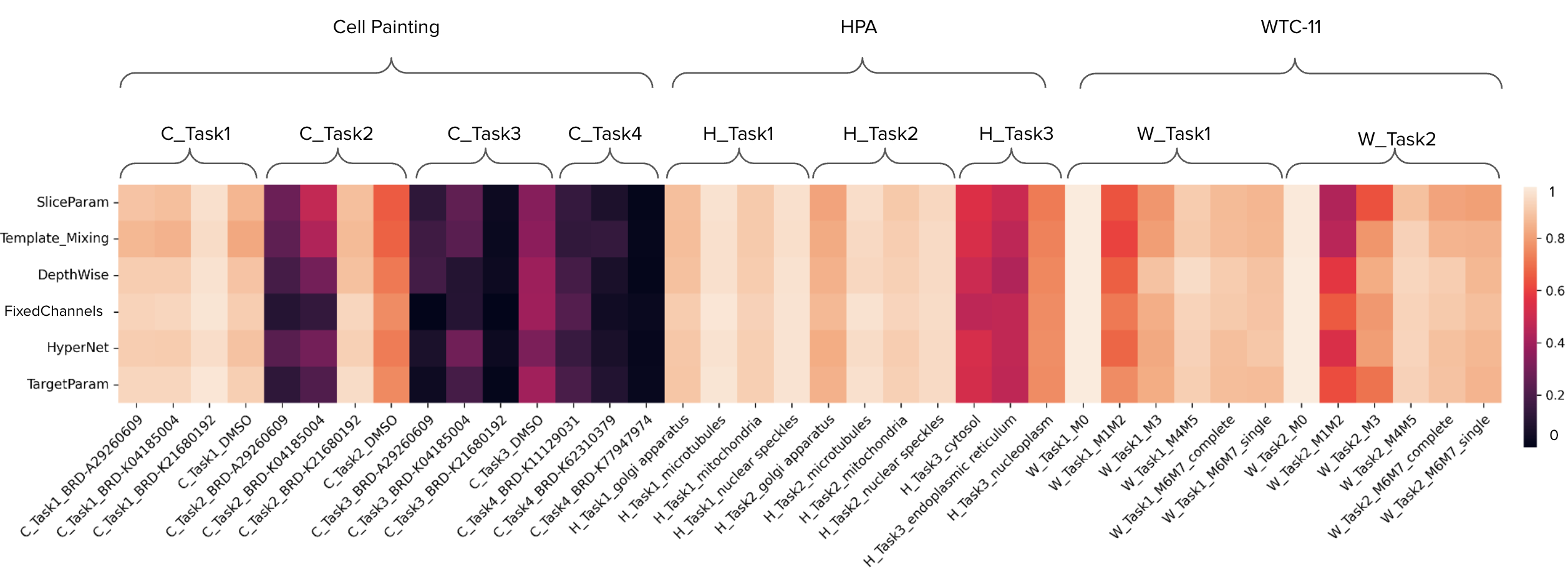}
  \vspace{-2mm}
  \caption{Heatmap of label-wise F1 score for each method. FixedChannels is the baseline model consisting of three models trained separately on the three subsets of CHAMMI. The other models are channel adaptive models trained on the combined training set of CHAMMI. All the models are pre-trained on ImageNet 22K and finetuned on the CHAMMI training set. The task labels on the x-axis follow the nomenclature of ‘Task\_Label’, where ‘Task’ is the task name (e.g. C\_Task1) and ‘Label’ is the classification label name (e.g. BRD-A29260609). 
}
  \label{fig:heatmap}
  \vspace{-2mm}
\end{figure}

\textbf{Class-level results} We present the detailed class-level F1 scores as a heatmap in Fig.~\ref{fig:heatmap}. On the y-axis, each row represents one model plotted in ascending order by the mean F1 score reported in Tab.~\ref{tab:full-f1}. The models include SliceParam, Template Mixing, Depthwise, FixedChannels (baseline), HyperNet, and TargetParam models. All the models included here are pre-trained and fine-tuned. On the x-axis, each column represents one class from one task. Lighter color indicates a higher F1 score and better performance for that class. We observe a pattern going from lighter to darker colors as we move from Task 1 to later tasks within each dataset, indicating increased difficulty. Interestingly, the M1M2 (prometaphase) class in WTC Task 1 and 2 shows notably lower performances compared to the other classes in WTC. This is likely due to the fact that prophase cells are morphologically similar to interphase cells (M0) and that class imbalance exists between M0 and M1M2 which we purposely preserve to resemble what we observe in real-life. Overall, compared to the baseline FixedChannels, TargetParam consistently shows similar and occasionally slightly better performances across classes.

\textbf{Domain generalization results } We evaluate the performance of baseline and channel adaptive models on out-of-distributions CHAMMI tasks when trained with existing domain generalization strategies SWAD and MIRO in Tab.~\ref{tab:ood-f1}. All the models are pre-trained on ImageNet 22K and fine-tuned on CHAMMI training set. Compared to training without domain generalization strategies, TargetParam, HyperNet, and Template mixing show improved overall performances with both SWAD and MIRO, while FixedChannels, Depthwise, SliceParam improve with SWAD but not MIRO. Channel-adaptive models also show comparable or better performances compared to the FixedChannels baseline when trained with domain generalization strategies. For instance, HyperNet trained with MIRO outperforms FixedChannels with MIRO in all but one task, and TargetParam trained with SWAD outperforms baseline in three out of six tasks.

\begin{table*}[t]
  \caption{Evaluation of existing generalization strategies SWAD~\cite{cha2021swad} and MIRO~\cite{cha2022domain} using fine-tuned baseline and channel-adaptive models. Average scores are calculated in the same way as in Tab.~\ref{tab:full-f1}}
  \label{tab:ood-f1}
  \setlength\tabcolsep{3pt}
  \centering
  \small
  \begin{tabular}{r*{10}{c}}
    \toprule
    & \multicolumn{4}{c}{Average OOD} 
    & \multicolumn{1}{c}{WTC} 
    & \multicolumn{2}{c}{HPA} 
    & \multicolumn{3}{c}{CP}   \\
    \cmidrule(r){2-5} \cmidrule(r){6-6} \cmidrule(r){7-8} \cmidrule(r){9-11}
    Model & Overall & WTC & HPA & CP & Task2 & Task2 & Task3 & Task2 & Task3 & Task4 \\
    \midrule
    FixedChannels   &  0.614&	0.861&	0.741&	0.240&	0.861&	\textbf{0.932}&	0.551&	0.484&	0.123	&0.112\\
    FixedChannels - MIRO & 0.612 & 0.819	& 0.737	& 0.281	& 0.819	& 0.916	& 0.559	& 0.528	& \textbf{0.226}	& 0.088 \\
    FixedChannels - SWAD & 0.628 & \textbf{0.877}	& 0.746	& 0.261	& \textbf{0.877}	& 0.931	& 0.561	& 0.484	& 0.177	& \textbf{0.121}\\
    \midrule
    Depthwise        & 0.616&	0.854&	0.737&0.258&	0.854	&0.918&	0.557	&0.520&	0.172	&0.081\\
    Depthwise - MIRO & 0.596 & 0.821	& 0.724	& 0.243	& 0.821	& 0.902	& 0.546	& 0.509	& 0.154	& 0.065 \\
    Depthwise - SWAD & 0.626 & 0.857	& 0.743	& 0.278	& 0.857	& 0.917	& 0.569	& 0.556	& 0.204	& 0.073 \\
    TargetParam      & 0.622	&0.843	&0.760	&0.264&	0.843	&0.925&	0.594	&0.512&	0.174	&0.107 \\
    TargetParam - MIRO & 0.625	& \textbf{0.877}	& 0.748	& 0.251	& \textbf{0.877}	& 0.928	& 0.567	& 0.533	& 0.131	& 0.088\\
    TargetParam - SWAD & 0.625	& 0.854	& 0.745	& 0.277	& 0.854	& 0.921	& 0.569	& 0.554	& 0.186	& 0.091\\
    SliceParam         & 0.592 &	0.751&	0.742&0.282&	0.751	&0.902&	0.582	&0.573&	0.202&	0.072 \\
    SliceParam - MIRO & 0.567	& 0.673	& 0.748	& 0.280	& 0.673	& 0.900	& \textbf{0.597}	& 0.545	& 0.202	& 0.094\\
    SliceParam - SWAD & 0.598	& 0.759	& 0.748	& 0.288	& 0.759	& 0.912	& 0.583	& 0.568	& 0.203	& 0.093\\
    HyperNet~\citep{haHypernetworks2016}& 0.616&	0.846	&0.745	&0.257	&0.846	&0.922	&0.568	&0.508	&0.170	&0.093\\
    HyperNet~\citep{haHypernetworks2016} - MIRO & 0.629	& 0.842	& \textbf{0.756}	& 0.289	& 0.842	& 0.922	& 0.590	& \textbf{0.583}	& 0.186	& 0.098\\
    HyperNet~\citep{haHypernetworks2016} - SWAD  & \textbf{0.632}	& 0.856	& 0.741	& \textbf{0.298}	& 0.856	& 0.922	& 0.560	& 0.582	& 0.192	& 0.119\\
    Template mixing~\citep{savarese2018learning} & 0.615 &0.823 &	0.743	 &0.279	 &0.823	 &0.919	 &0.566	 &0.542 &	0.202	 &0.094 \\
    Template mixing~\citep{savarese2018learning} - MIRO & 0.614	& 0.847	& 0.731	& 0.263	& 0.847	& 0.918	& 0.545	& 0.549	& 0.154	& 0.087\\
    Template mixing~\citep{savarese2018learning} - SWAD & 0.621	& 0.851	& 0.733	& 0.280	& 0.851	& 0.919	& 0.547	& 0.579	& 0.195	& 0.067\\
    \bottomrule
  \end{tabular}
  \vspace{-2mm}
\end{table*}

\begin{table*}[t]
  \caption{Evaluation of TPS~\cite{tang2019augmentation} transformation and SimCLR~\cite{chen2020simple,chen2020big} self-supervised learning (SSL) framework using fine-tuned baseline and channel-adaptive models. See Section B.3 for more details about TPS and the loss functions. \enquote{Baseline} condition: model used supervised ProxyNCA++ loss ($L_{\text{ProxyNCA++}}$) and standard data transformation (without TPS). \enquote{TPS} and \enquote{SSL}:  model trained with either TPS or $L_{\text{SSL}}$ only. \enquote{TPS + SSL}: model trained with both TPS and $L_{\text{SSL}}$. \enquote{TPS + 0.2 SSL}: model trained with TPS and $L_{\text{combined}}$, which consists of 0.2 SSL loss and 0.8 ProxyNCA++ loss. Average scores are calculated in the same way as in Tabs.~\ref{tab:full-f1} and \ref{tab:ood-f1}}.
  \label{tab:tps&simclr}
  \centering
  \small
  \setlength\tabcolsep{2.5pt}
  \begin{tabular}{r*{11}{c}}
    \toprule
    & \multicolumn{5}{c}{Overall Score} 
    & \multicolumn{1}{c}{WTC} 
    & \multicolumn{2}{c}{HPA} 
    & \multicolumn{3}{c}{CP}   \\
    \cmidrule(r){3-6} \cmidrule(r){7-7} \cmidrule(r){8-9} \cmidrule(r){10-12}
    Model & Condition & Overall & WTC & HPA & CP & Task2 & Task2 & Task3 & Task2 & Task3 & Task4 \\
    \midrule
    FixedChannels & Baseline & 0.614&	0.861&	0.741&	0.240&	0.861&	0.932&	0.551&	0.484&	0.123	&0.112\\
    FixedChannels & SSL & 0.368&	0.385&	0.457&	\textbf{0.260} &	0.385	&0.594	&0.320	&0.447	&0.227	&0.107 \\
    FixedChannels & TPS & 0.623	&0.876	&0.748	&0.245&	0.876&	0.932&	0.564	&0.494&	0.133	&0.109\\
    FixedChannels & TPS + SSL & 0.359	&0.398	&0.425	&0.255	&0.398&	0.540	&0.309&	0.426	&0.217	&0.123\\
    FixedChannels & TPS + 0.2 SSL & \textbf{0.633} &	\textbf{0.879}&	\textbf{0.766}	&0.253	&0.879&	0.920&	0.613&	0.499&	0.164&	0.097\\
    \midrule
    Depthwise & Baseline & 0.616&	0.854&	0.737&	\textbf{0.258}&	0.854	&0.918&	0.557	&0.520&	0.172	&0.081\\
    Depthwise & SSL & 0.205&	0.185	&0.231	&0.198	&0.185	&0.261	&0.202&	0.280&	0.221	&0.094\\
    Depthwise & TPS& \textbf{0.622}	&\textbf{0.877}	&\textbf{0.755}	&0.235	&0.877&	0.920	&0.591&	0.496	&0.113&	0.097 \\
    Depthwise & TPS + SSL& 0.204	&0.176&	0.236	&0.199	&0.176&	0.270&	0.203&	0.266	&0.228	&0.104 \\
    Depthwise & TPS + 0.2 SSL& 0.608&	0.818&	0.754&	0.251&	0.818	&0.917&	0.592	&0.506	&0.170&	0.078\\
    \cmidrule(r){1-12}
    SliceParam & Baseline & 0.592&	0.751&	0.742&	\textbf{0.282}&	0.751	&0.902&	0.582	&0.573&	0.202&	0.072\\
    SliceParam & SSL & 0.259&	0.267	&0.283	&0.226	&0.267&	0.341	&0.226&	0.335	&0.251	&0.093\\
    SliceParam & TPS & \textbf{0.600} &	\textbf{0.775}&	0.751&	0.275	&0.775&	0.913&	0.588	&0.574&	0.177&	0.075\\
    SliceParam & TPS + SSL& 0.264&	0.274	&0.291&	0.227&	0.274	&0.339&	0.244	&0.344	&0.243&	0.094 \\
    SliceParam & TPS + 0.2 SSL & \textbf{0.600}	&0.758	&\textbf{0.767}	&0.276	&0.758	&0.901&	0.633	&0.544	&0.210	&0.074\\
    \cmidrule(r){1-12}
    TargetParam & Baseline & 0.622	&\textbf{0.843}	&0.760	&0.264&	0.843	&0.925&	0.594	&0.512&	0.174	&0.107\\
    TargetParam & SSL & 0.247&	0.194&	0.307&	0.239	&0.194	&0.375	&0.240	&0.351	&0.256&	0.109\\
    TargetParam & TPS & \textbf{0.628}	&0.842	&\textbf{0.779} &	0.262	&0.842	&0.935	&0.623	&0.513	&0.167	&0.105\\
    TargetParam & TPS + SSL & 0.241&	0.201&	0.297	&0.225	&0.201	&0.363	&0.231&	0.334&	0.242&	0.100\\
    TargetParam & TPS + 0.2 SSL& 0.618&	0.822&	0.759	&\textbf{0.275} &	0.822	&0.935&	0.583&	0.564&	0.161&	0.100\\
    \cmidrule(r){1-12}
    Template mixing & Baseline & 0.615 &	\textbf{0.823} &	0.743	 &0.279	 &0.823	 &0.919	 &0.566	 &0.542 &	0.202	 &0.094 \\
    Template mixing & SSL &0.230&	0.224	&0.251&	0.215	&0.224	&0.291& 0.211&	0.313&	0.239&	0.093\\
    Template mixing & TPS & 0.616&	0.818	&\textbf{0.760} &	0.270&	0.818	&0.910&	0.610&	0.536	&0.189	&0.084\\
    Template mixing & TPS + SSL &0.237&	0.253&	0.248&	0.210&	0.253	&0.284&	0.212&	0.301	&0.237	&0.092\\
    Template mixing & TPS + 0.2 SSL& \textbf{0.621}	&0.822	&0.754	&\textbf{0.286}	&0.822&	0.899	&0.610	&0.551&	0.215	&0.090\\
    \cmidrule(r){1-12}
    HyperNet & Baseline & 0.616&	0.846	&0.745	&0.257	&0.846	&0.922	&0.568	&0.508	&0.170	&0.093\\
    HyperNet & SSL & 0.200	&0.174	&0.225	&0.202&	0.174&	0.255	&0.195	&0.266&	0.233&	0.106\\
    HyperNet & TPS & 0.630	&0.854	&\textbf{0.761}	&0.276	&0.854	&0.926	&0.595	&0.556	&0.156	&0.118\\
    HyperNet & TPS + SSL & 0.202	&0.172	&0.232	&0.202	&0.172	&0.273	&0.191	&0.260	&0.241	&0.105\\
    HyperNet & TPS + 0.2 SSL & \textbf{0.632} &	\textbf{0.855} &	0.756 &	\textbf{0.283} &	0.855	 &0.923	 &0.589	 &0.563	 &0.189	 &0.097\\
    
    \bottomrule
  \end{tabular}
  \vspace{-2mm}
\end{table*}

\textbf{Data augmentation and loss functions } We evaluate the TPS~\cite{tang2019augmentation} transformation in comparison to baseline random cropping and horizontal flips. Results in Tab.~~\ref{tab:tps&simclr} indicate that TPS increased the overall F1 score of all models (comparing \enquote{TPS} to \enquote{Baseline} within each model). As mentioned in Tab.~1 of the main paper, the increase is up to 2\% for HyperNet. Interestingly, using TPS decreases the F1 score of CP dataset for channel-adaptive models but not for the FixedChannel model.

In addition, we evaluated self-supervised learning (SSL) for solving the tasks in CHAMMI. Compared to the baseline performance, using $L_{\text{SSL}}$ alone as the loss function severely undermines the performances (\enquote{SSL} vs. \enquote{Baseline}, Tab.~\ref{tab:tps&simclr}) of all models. When we combine TPS and $L_{\text{SSL}}$, the models continue to underperform. And finally, when we combine the supervised loss function $L_{\text{ProxyNCA++}}$ and $L_{\text{SSL}}$, as well as using the TPS transformation, the performance increased slightly for FixedChannels, Template mixing, and HyperNet models, while staying the same or decreased for the other models. 



\end{document}